%% file: PID6017-arxiv.tex
\documentclass[letterpaper]{article}
\usepackage[preprint]{aaai2027}
\usepackage[hyphens]{url}
\usepackage{graphicx}
\usepackage{natbib}
\usepackage{caption}
\usepackage{amsmath,amssymb}
\usepackage{algorithm}
\usepackage{algorithmic}
\usepackage{booktabs}
\usepackage{multirow}
\usepackage{colortbl}
\usepackage{xspace}

\definecolor{rankfirst}{HTML}{F7C7CE}
\definecolor{ranksecond}{HTML}{FFF1B8}
\definecolor{rankthird}{HTML}{DDEFD7}
\definecolor{revisionblue}{RGB}{0,0,255}
\newcommand{\firstcell}[1]{\cellcolor{rankfirst}\textbf{#1}}
\newcommand{\secondcell}[1]{\cellcolor{ranksecond}\underline{#1}}
\newcommand{\thirdcell}[1]{\cellcolor{rankthird}#1}


\newcommand{\method}{MOSAIC\xspace}
\newcommand{\feddda}{FedDDA\xspace}
\newcommand{\R}{\mathbb{R}}
\newcommand{\pdfanchor}[1]{\pdfdest name {#1} xyz}
\newcommand{\pdfjump}[2]{\leavevmode\pdfstartlink attr{/Border [0 0 0]} goto name {#1}#2\pdfendlink}
\newcommand{\figref}[1]{\pdfjump{#1}{Figure~\ref{#1}}}
\newcommand{\tabref}[1]{\pdfjump{#1}{Table~\ref{#1}}}

\usepackage{array}

\title{Beyond Domain-Level Adaptation: Margin-Oriented Semantic-Appearance Interaction Correction for Personalized Federated Vision-Language Models}
\author{
Wentao Yue, Qingyu Mao, Tianyou Lai,\\
Ahmed M. Abdelmoniem, Qilei Li
}
\affiliations{}

\begin{document}

\maketitle

\begin{abstract}
Federated parameter-efficient fine-tuning enables distributed clients to adapt pretrained vision-language models without sharing raw data or updating the full backbone. Its effectiveness, however, is limited by domain heterogeneity across clients. Existing personalized methods separate globally shared knowledge from client-specific style, but they largely treat each domain as a class-agnostic transformation. We show that this abstraction is insufficient: the cross-domain displacement associated with a fixed domain varies across semantic classes, and only a subset of these class-domain residuals damages the image-text decision margin. We therefore propose Margin-Oriented Semantic-Appearance Interaction Correction (\method), which first constructs a decision-aware harmfulness score that measures whether a training-derived class-domain residual favors a competing text prototype over the true class. It then models fine-grained class-domain interactions with a low-rank residual adapter whose class factors and residual basis are globally shared while domain factors remain client-private. An image-conditioned gate further controls candidate-wise correction, and harmful-pair-aware reweighting prioritizes decision-relevant residuals during local optimization. Extensive experiments on Office31, OfficeHome, and DomainNet100 demonstrate that \method consistently improves macro-client top-1 accuracy across all evaluated domain-shift and joint domain-label-shift settings.
\end{abstract}

\section{Introduction}

Pretrained vision-language models (VLMs), such as Contrastive Language--Image Pre-training (CLIP)~\citep{radford2021clip}, acquire transferable visual concepts through large-scale image-text alignment. The zero-shot CLIP classifier uses this alignment without downstream adaptation and provides an unadapted reference for prompt-based methods. Parameter-efficient fine-tuning (PEFT) adapts this knowledge to downstream tasks by updating only compact prompts or adapters while keeping the backbone frozen. Prompt-learning methods, including CoOp~\citep{zhou2022coop} and its conditional extension CoCoOp~\citep{zhou2022cocoop}, demonstrate that a small number of learned context tokens can effectively adapt a frozen VLM. Combining PEFT with federated learning (FL) yields an efficient and privacy-conscious paradigm in which distributed clients collaboratively adapt a VLM without exchanging their raw images. This combination is particularly attractive for multi-source visual recognition, where data ownership and communication cost prevent centralized training.

\begin{figure}[t]
  \centering
  \includegraphics[width=1\columnwidth]{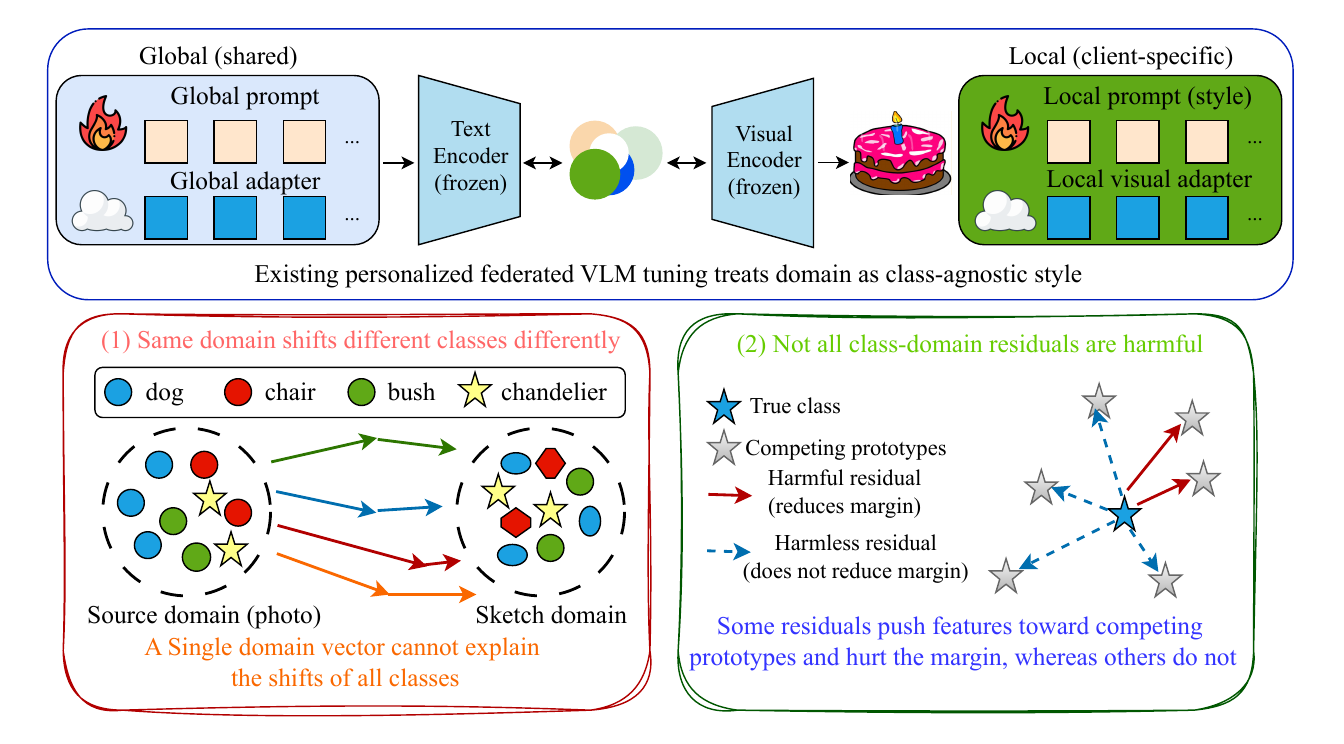}
  \caption{
  Motivation of MOSAIC. Existing personalized federated VLM tuning usually treats each domain as a class-agnostic style, while the same domain can shift different semantic classes differently and only some class-domain residuals harm the decision margin.
  }
  \pdfanchor{fig:motivation}\label{fig:motivation}
\end{figure}

The main obstacle is client heterogeneity. Images collected by different clients may differ in acquisition conditions, rendering conventions, and background statistics, producing domain-dependent feature distributions~\citep{yue2026review,li2026federated}. Classical federated optimization addresses this mismatch with a shared model (FedAvg~\citep{mcmahan2017fedavg}) or by constraining local updates around the global solution (FedProx~\citep{li2020fedprox}). Personalized FL instead preserves client-specific parameters, for example local normalization statistics in FedBN~\citep{li2021fedbn} or private prediction heads in FedRep~\citep{collins2021fedrep}. These approaches are valuable foundations, but they do not specify how a frozen image-text representation should distinguish transferable semantic structure from domain-dependent visual variation. Personalized federated VLM fine-tuning mitigates this problem by separating globally shared knowledge from client-private style. For example, \feddda~\citep{yang2025feddda} decouples global and local textual prompts and combines shared and client-specific visual adapters through a dynamic gate. However, this family of designs makes a coarse modeling choice: it treats a domain as a client-level, largely class-agnostic transformation.

\begin{figure*}[!t]
    \centering
    \includegraphics[width=0.95\textwidth]{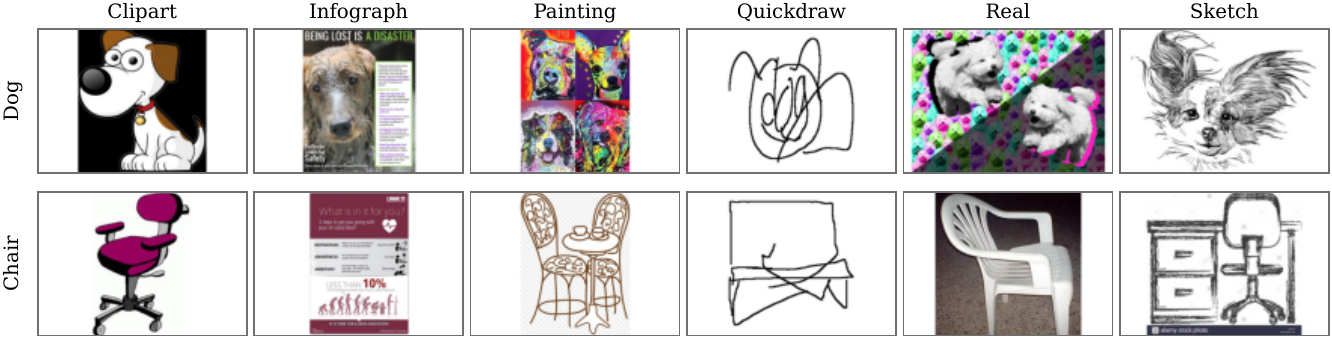}
    \caption{Class-conditioned domain variation on DomainNet100. Rows show dog and chair, while columns cover all six rendering domains. Cross-domain changes differ between the two semantic classes, indicating that domain heterogeneity cannot be represented faithfully by a single class-agnostic shift.}
    \pdfanchor{fig:class-conditioned-drift}\label{fig:class-conditioned-drift}
\end{figure*}

Our starting observation is that this abstraction leaves a structured residual. A domain does not act identically on every semantic class: a sketch changes a dog's fur, pose, and curved contours differently from how it changes a chair's geometry, straight edges, and viewpoint. Thus, after accounting for class semantics and a common domain effect, the displacement of class $c$ in domain $d$ still contains a class-conditioned domain residual. \figref{fig:motivation} summarizes the resulting gap: the upper panel shows the class-agnostic shared/private decomposition used by existing personalized VLM tuning, the lower-left panel illustrates class-dependent visual changes within one domain, and the lower-right panel distinguishes residuals that reduce the true-versus-competitor margin from harmless ones. \figref{fig:class-conditioned-drift} expands this observation to all six DomainNet rendering domains using dog and chair as representative classes. Across a row, one class shifts differently across domains; down a column, one domain shifts dog and chair differently. A class-agnostic domain vector would produce aligned class shifts, but these different directions reveal a class-domain interaction.

\begin{figure*}[!t]
    \centering
    \includegraphics[width=0.88\textwidth]{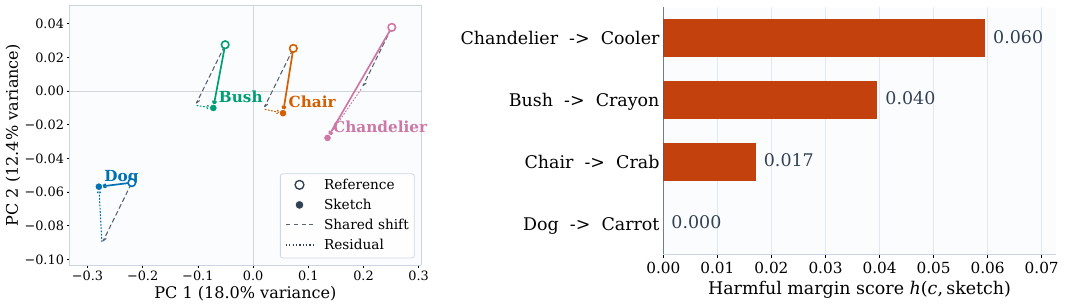}
    \caption{Feature-space evidence on DomainNet100. \textbf{Left:} class-specific shifts into sketch deviate from one shared shift. \textbf{Right:} only residuals that reduce the true-versus-competitor margin receive positive harmfulness scores. The harmful margin score is defined in Eq.~\ref{eq:harmful-score}.}
    \pdfanchor{fig:feature-evidence}\label{fig:feature-evidence}
\end{figure*}

Feature-space evidence makes the diagnosis quantitative. In the left panel of \figref{fig:feature-evidence}, the shifts of four classes into the sketch domain point in different directions, and one shared shift cannot reconstruct them. This verifies that the residual is not merely a visual anecdote. Yet a geometric residual alone does not specify what to correct: the dog residual is visible but does not damage classification, whereas the residuals of chair, bush, and chandelier move their features toward competing text prototypes. We therefore score each class-domain residual by how strongly it favors the most competitive incorrect text prototype over the true class; the exact definition is given in Eq.~\ref{eq:harmful-score}. The right panel yields our second key finding: only decision-harmful class-domain residuals should be corrected. Principal component analysis (PCA) is used only for visualization; all residual and margin calculations use the original VLM feature space.

These two findings expose three requirements. First, a correction must represent the interaction between semantic class and local domain rather than adding one domain vector to every class. Second, it must be decision-aware, because residual norm is not equivalent to classification harm. Third, in a federated system it must preserve transferable class structure without averaging away client-private domain effects, and it must avoid perturbing samples whose residuals are harmless.

To meet these requirements, we propose \method, a decision-aware residual framework for personalized federated VLM tuning. Its factorized adapter represents class-domain residuals through a shared low-rank basis and compact class and private-domain coefficients, avoiding an independent high-dimensional vector for every pair. Its harmfulness score measures the change in the true-versus-competitor text margin and drives harmful-pair-aware local optimization. Finally, \method aggregates only shared class-side structure and uses an image-conditioned, candidate-wise gate to subtract a residual only when it is useful. The design therefore turns the problem discovery into a targeted correction mechanism while preserving the parameter efficiency of compact textual and visual adaptation.

\textbf{Our contributions are summarized as follows:}

\begin{itemize}
    \item We identify harmful class-conditioned domain drift as an overlooked source of degradation in personalized federated VLM fine-tuning, and formulate it as the class-domain interaction that remains beyond class semantics and a common domain effect.
    \item We introduce a decision-aware harmfulness score that measures whether a class-domain residual reduces the image-text classification margin, avoiding the unreliable assumption that larger residuals are more harmful.
    \item We develop \method, a low-rank factorized residual adapter with globally shared class structure, client-private domain factors, image-conditioned candidate-wise correction, and harmful-pair-aware local optimization.
\end{itemize}

\section{Related Work}

\paragraph{Parameter-Efficient VLM Tuning.}
Parameter-efficient VLM adaptation freezes the foundation model and learns a small prompt or adapter state. CoOp~\citep{zhou2022coop} and CoCoOp~\citep{zhou2022cocoop} established continuous and image-conditioned textual contexts, respectively. Recent work focuses on retaining the generalization of the frozen VLM while adapting it: Knowledge-Guided Context Optimization (KgCoOp)~\citep{yao2023kgcoop} constrains learned prompts with hand-crafted textual knowledge, PromptSRC~\citep{khattak2023promptsrc} self-regularizes prompt trajectories, ProMetaR~\citep{park2024prometar} meta-learns prompt regularization, and Textual-based Class-aware Prompt Tuning (TCP)~\citep{yao2024tcp} makes textual contexts class-aware. These methods commonly use prompt-side regularization or conditioning to improve transfer with a compact trainable state. However, they model adaptation at the task, image, or class-prompt level; they do not identify the residual induced jointly by a semantic class and a decentralized visual domain, nor do they determine whether that residual harms a particular image-text margin. \method retains the PEFT efficiency principle but performs correction in the aligned feature space only for decision-harmful class-domain interactions.

\paragraph{Personalized Federated VLM Tuning.}

Federated VLM tuning extends compact adaptation to private non-IID clients. PromptFL~\citep{guo2023promptfl} aggregates soft prompts, while PromptFL+Prox adds a proximal penalty; FedTPG~\citep{qiu2024fedtpg} generates prompts for new classes. FedOTP~\citep{li2024fedotp}, FedPGP~\citep{cui2024fedpgp}, and DiPrompT~\citep{bai2024diprompt} balance shared and personalized prompt knowledge. FedPHA~\citep{fang2025fedpha} and FOCoOp~\citep{liao2025focoop} address heterogeneous prompt capacity or OOD robustness, whereas \feddda~\citep{yang2025feddda} decouples textual priors and dynamically adapts visual features. These methods specialize prompts or adapters at the client or domain level; none represents a class-specific domain residual and tests its effect on the true-versus-competitor margin.

\paragraph{Domain Shift and Fine-Grained Interactions.}
Federated domain generalization studies how to learn transferable representations when source domains are isolated across clients. Earlier work sought domain invariance through adversarial alignment and covariance matching~\citep{ganin2016dann,sun2016deepcoral}, representation regularization~\citep{nguyen2022fedsr}, or class-prototype exchange~\citep{tan2022fedproto}. Generalization Adjustment~\citep{zhang2023fedga} calibrates aggregation weights to compensate for the absence of joint multi-domain batches, and FedOMG~\citep{nguyen2025fedomg} matches gradients on the server without sharing client data. Recent foundation-model methods instead introduce domain-specific adaptation states: FedAG~\citep{wang2025fedag} uses multiple fine-grained adapters to integrate domain knowledge, and FedDSPG~\citep{wu2025feddspg} generates domain-specific soft prompts for unseen targets. Together, these methods exploit aggregation, prompts, or adapters to reduce domain sensitivity and improve cross-domain transfer. Yet their correction target is still a domain-invariant representation, a domain-level prompt, or an adapter mixture. They do not separate the common effect of a domain from the class-conditioned residual that remains after it, and they do not use the residual's effect on a competing text prototype to decide whether intervention is warranted. \method makes these two distinctions explicit, linking fine-grained diagnosis to a selective margin-oriented correction during personalized federated optimization.

\begin{figure*}[!t]
    \centering
    \includegraphics[width=0.98\textwidth]{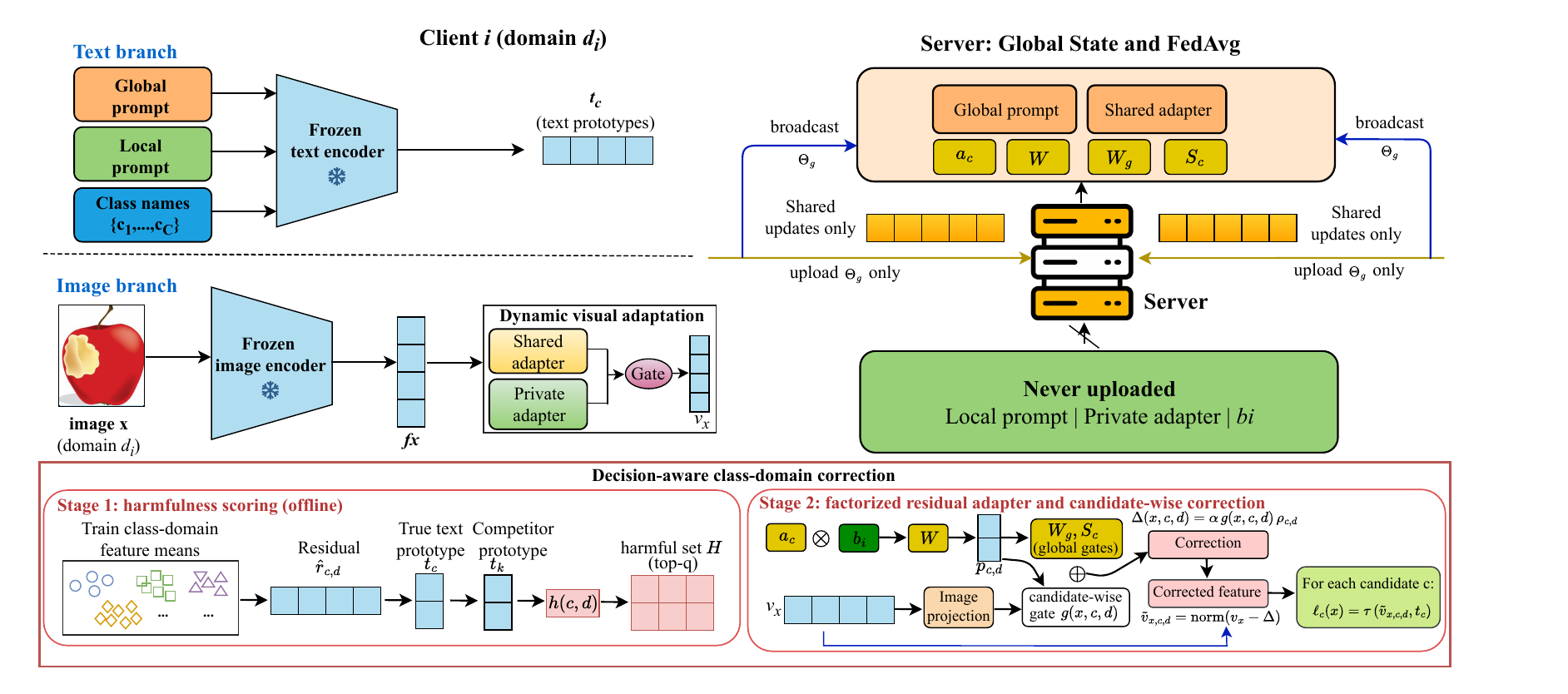}
    \caption{Overview of \method. Stage 1 identifies harmful class-domain pairs; Stage 2 factorizes and gates candidate-wise residual correction. Only shared parameters are aggregated.}
    \pdfanchor{fig:mosaic-overview}\label{fig:mosaic-overview}
\end{figure*}

\section{Problem Formulation}
Our goal is to correct class-conditioned domain effects without averaging them away across clients. This requires distinguishing what should transfer across the federation from what must remain local to a visual domain. We therefore formulate \method as personalized VLM adaptation with an explicitly partitioned shared/private state.

We consider $M$ clients and a central server. Client $i$ owns a private dataset $\mathcal{D}_i=\{(x_j,y_j,d_i)\}_{j=1}^{n_i}$ from domain $d_i$. Client $i$ belongs to domain $d_i$ and retains a private domain factor $b_i\in\R^r$; the client index emphasizes that this factor remains local even when multiple clients belong to the same visual domain. \method adapts a frozen VLM with image and text encoders $F_I$ and $F_T$ through compact shared and private states. Its shared state $\Theta^g$ contains global prompt and visual-adapter parameters together with the class-side residual parameters; its client-private state $\Theta_i^p$ contains private prompt and visual-adapter parameters and $b_i$. Having separated these roles, the server aggregates only the shared state at communication round $t$:
\begin{equation}
\theta_{t+1}^{g}=\sum_{i\in\mathcal{C}_t}\frac{n_i}{\sum_{j\in\mathcal{C}_t}n_j}\theta_{i,t}^{g},
\label{eq:fed-aggregation}
\end{equation}
where $\mathcal{C}_t$ denotes the participating clients.

For an image $x$, let $v_x\in\R^D$ and $t_c\in\R^D$ be the normalized visual and textual features produced by the frozen encoders conditioned on MOSAIC's shared and client-private adaptation states. Before residual correction, standard VLM classification uses
\begin{equation}
p(y=c\mid x)=\frac{\exp(\tau\langle v_x,t_c\rangle)}{\sum_{k=1}^{C}\exp(\tau\langle v_x,t_k\rangle)},
\label{eq:base-classification}
\end{equation}
where $\tau$ is the learned logit scale.

\section{Methodology}

\figref{fig:mosaic-overview} summarizes two stages built on the shared/private federated VLM defined above. Stage 1 runs once before federated optimization. Training-only class-domain feature statistics remove the displacement shared across classes, producing an interaction residual for each pair. The most adverse competitor-versus-true projection of each residual defines its harmfulness score and a fixed harmful set $\mathcal{H}$. Stage 2 learns each candidate residual as $\rho_{c,i}=W(a_c\odot b_i)$: $a_c$ and $W$ capture globally transferable class structure, whereas $b_i$ remains client-private. An image-conditioned gate controls the correction separately for every candidate class. Local training upweights harmful pairs, while the server aggregates only shared parameters.

\subsection{Class-Conditioned Domain Drift}

A domain-level shift alone cannot describe the observed mismatch: the same domain changes the appearance of different semantic classes in different directions. A useful correction target must therefore remove the domain effect shared by all classes and retain only the class-specific remainder. Let $v_x^{\mathrm{diag}}\in\R^D$ denote the normalized visual feature extracted from training image $x$ by the fixed diagnostic model, and let $N_{c,d}$ be the number of training images for class $c$ in domain $d$. We define the empirical class-domain centroid as $\bar v_{c,d}=N_{c,d}^{-1}\sum_{j:y_j=c,d_j=d}v_{x_j}^{\mathrm{diag}}$, where the sum uses training samples only. The diagnostic checkpoint, deterministic sampling protocol, and handling of missing pairs are detailed in the supplementary material. A class-agnostic model approximates this centroid as
\begin{equation}
\bar v_{c,d}\approx m_c+u_d,
\end{equation}
where $m_c$ represents class semantics and $u_d$ is a domain effect shared across classes. We augment this restricted model with a class-domain interaction:
\begin{equation}
\bar v_{c,d}=m_c+u_d+r_{c,d}+\epsilon_{c,d},
\end{equation}
where $r_{c,d}$ captures the class-domain interaction that cannot be explained by a common domain vector.

We estimate this interaction using training features only. First, a leave-one-domain-out class displacement is
\begin{equation}
\delta_{c,d}=\bar v_{c,d}-\bar v_{c,-d}, \qquad
\bar v_{c,-d}=\frac{1}{|\mathcal{D}|-1}\sum_{d'\neq d}\bar v_{c,d'}.
\end{equation}
We then remove the displacement shared by other classes in domain $d$:
\begin{equation}
\widehat r_{c,d}=\delta_{c,d}-\frac{1}{C-1}\sum_{c'\neq c}\delta_{c',d}.
\label{eq:interaction-residual}
\end{equation}
Equation~\ref{eq:interaction-residual} isolates the additional effect of domain $d$ on class $c$ beyond the domain's common action.

\subsection{Decision-Aware Harmful Drift Scoring}
The residual in Eq.~\ref{eq:interaction-residual} is geometric, whereas classification errors are caused by an unfavorable ranking against a competing text prototype. Consequently, correcting every large residual would perturb samples whose decisions are already safe. \method instead asks whether a residual specifically favors a competing text prototype over the true-class prototype. For a competitor $k\neq c$, its margin damage is
\begin{equation}
q_k(c,d)=\langle t_k,\widehat r_{c,d}\rangle-\langle t_c,\widehat r_{c,d}\rangle.
\label{eq:margin-damage}
\end{equation}
The harmfulness score is
\begin{equation}
h(c,d)=\max\left(0,\max_{k\neq c}q_k(c,d)\right).
\label{eq:harmful-score}
\end{equation}
A positive score indicates that the residual points more strongly toward at least one competing text class than toward the true class. Within each domain, we rank class-domain pairs by $h(c,d)$ and mark the top quantile as the harmful set $\mathcal{H}$. This construction prevents domains with larger feature variation from dominating the selected pairs.

\subsection{Factorized Class-Domain Residual Adapter}
After identifying a harmful pair, the model must represent its residual without allocating a separate high-dimensional vector to every class-domain combination. Such independent vectors are costly and would prevent clients from sharing semantic structure. \method therefore learns a global class factor $a_c\in\R^r$, the client-private domain factor $b_i\in\R^r$, and a global basis $W\in\R^{D\times r}$. Their interaction is
\begin{equation}
z_{c,i}=a_c\odot b_i, \qquad \rho_{c,i}=Wz_{c,i},
\label{eq:factorized-residual}
\end{equation}
where $r\ll D$ and $\odot$ denotes element-wise multiplication. The factorization allows clients to share how semantic classes use a common residual basis while retaining how their local domain activates that basis.

\subsection{Image-Conditioned Candidate-Wise Correction}
Even within a harmful class-domain pair, residual severity varies across images and across candidate labels. Applying one fixed vector would therefore over-correct easy examples and cannot target the actual competitor. \method uses an image-conditioned, candidate-wise gate to determine how strongly each candidate residual should act. It projects the image feature through $W_g\in\R^{r\times D}$ and computes
\begin{equation}
g(x,c,i)=\tanh(s_c)\,\sigma\left(\frac{\langle W_gv_x,z_{c,i}\rangle}{\sqrt r}\right),
\label{eq:candidate-gate}
\end{equation}
where $s_c$ is a learnable class amplitude. The candidate-specific correction is
\begin{equation}
\Delta(x,c,i)=\alpha\,g(x,c,i)\,\rho_{c,i},
\label{eq:candidate-correction}
\end{equation}
with residual scale $\alpha$. Because $\rho_{c,i}$ models an undesirable displacement, we subtract it and normalize the result:
\begin{equation}
\widetilde v_{x,c,i}=\frac{v_x-\Delta(x,c,i)}{\|v_x-\Delta(x,c,i)\|_2}.
\label{eq:corrected-feature}
\end{equation}
Each candidate class obtains its own corrected visual feature, yielding logits
\begin{equation}
\ell_c(x)=\tau\langle\widetilde v_{x,c,i},t_c\rangle.
\label{eq:corrected-logit}
\end{equation}
The learned gate starts from a near-identity correction, preserving the pretrained representation early in training.

\subsection{Harmful-Pair-Aware Federated Optimization}
The harmfulness score identifies where correction is needed, but the local objective must also make these rare, decision-relevant pairs influential during optimization. We therefore upweight only the samples from the selected harmful set while retaining ordinary cross-entropy learning for all other data. For sample $(x_j,y_j,d_i)$ at client $i$, we assign
\begin{equation}
w_j=1+(\gamma-1)\mathbb{I}[(y_j,d_i)\in\mathcal{H}],
\label{eq:sample-weight}
\end{equation}
where $\gamma\geq1$ controls harmful-pair emphasis. The local objective is
\begin{equation}
\mathcal{L}_i=
\frac{\sum_jw_j\operatorname{CE}(\ell(x_j),y_j)}{\sum_jw_j}
+\lambda\mathcal{R}_{\mathrm{res}},
\label{eq:local-objective}
\end{equation}
where $\operatorname{CE}$ denotes cross-entropy and $\mathcal{R}_{\mathrm{res}}$ regularizes the class factors, private domain factors, global basis, and image gate.

The server aggregates the class factors $\{a_c\}$, residual basis $W$, image projection $W_g$, class amplitudes $\{s_c\}$, and MOSAIC's global prompt and visual-adapter parameters according to Eq.~\ref{eq:fed-aggregation}. Each client retains its domain factor $b_i$ together with its private prompt and visual-adapter parameters. Algorithm~\ref{alg:hcdda} summarizes the optimization and explicitly follows the harmful-pair construction in Eq.~\ref{eq:harmful-score}, the candidate-wise correction in Eqs.~\ref{eq:candidate-gate}--\ref{eq:corrected-logit}, and the local objective in Eq.~\ref{eq:local-objective}.

\begin{algorithm}[t]
\caption{Personalized Federated Optimization of \method}
\label{alg:hcdda}
\begin{algorithmic}[1]
\REQUIRE Clients $\{\mathcal{D}_i\}_{i=1}^M$, rounds $T$, harmful pairs $\mathcal{H}$ from Eq.~\ref{eq:harmful-score}, shared parameters $\Theta^g$, private parameters $\{\Theta_i^p\}$
\FOR{$t=0,\ldots,T-1$}
    \STATE Server broadcasts $\Theta_t^g$ to participating clients $\mathcal{C}_t$
    \FOR{each client $i\in\mathcal{C}_t$ in parallel}
        \STATE Load $\Theta_t^g$ and retain private domain factor $b_i$
        \FOR{each local minibatch $\mathcal{B}\subset\mathcal{D}_i$}
            \STATE Compute MOSAIC visual features $v_x$ and text features $t_c$
            \STATE Generate $\rho_{c,i}$ using Eq.~\ref{eq:factorized-residual}
            \STATE Compute $g$, $\Delta$, and $\ell_c(x)$ using Eqs.~\ref{eq:candidate-gate}--\ref{eq:corrected-logit}
            \STATE Update shared and private parameters using Eq.~\ref{eq:local-objective}
        \ENDFOR
        \STATE Upload only the shared parameter update $\Theta_{i,t+1}^g$
    \ENDFOR
    \STATE Aggregate $\Theta_{t+1}^g$ according to Eq.~\ref{eq:fed-aggregation}
\ENDFOR
\RETURN Shared parameters $\Theta_T^g$ and personalized parameters $\{\Theta_i^p\}$
\end{algorithmic}
\end{algorithm}

\begin{table*}[!t]
\centering
\setlength{\tabcolsep}{3.5pt}
\renewcommand{\arraystretch}{1.08}
\resizebox{0.96\textwidth}{!}{%
\begin{tabular}{lcccc|ccccc|ccccccc}
\toprule
\multirow[c]{2}{*}{Method} & \multicolumn{4}{c}{Office31} & \multicolumn{5}{c}{OfficeHome} & \multicolumn{7}{c}{DomainNet100} \\
& A & W & D & Avg. & A & C & P & R & Avg. & C & I & P & Q & R & S & Avg. \\
\midrule
\multicolumn{17}{l}{\textit{Setting \textcircled{1}: one domain for one client}} \\
Zero-shot CLIP [ICML2021] & 81.14 & 72.45 & 74.05 & 75.88 & 84.30 & 66.28 & 89.06 & 89.66 & 82.33 & 71.93 & 53.30 & 65.73 & 13.57 & 83.49 & 66.46 & 59.08 \\
PromptFL [TMC2023] & 88.90 & 87.55 & 94.30 & 90.25 & 86.94 & 75.76 & 94.32 & \secondcell{93.59} & 87.65 & 86.55 & 70.29 & 79.89 & 34.31 & 91.54 & 79.97 & 73.76 \\
PromptFL+Prox [TMC2023; MLSys2020] & \thirdcell{89.22} & 89.80 & 93.04 & 90.68 & 86.16 & 76.28 & 94.25 & \secondcell{93.59} & 87.57 & 87.47 & 71.25 & 82.15 & 32.63 & 91.79 & 81.20 & 74.41 \\
FedOTP [CVPR2024] & 85.73 & 94.69 & 94.94 & 91.79 & 79.71 & 76.24 & 92.18 & 87.10 & 83.81 & 86.73 & 69.80 & 82.05 & 50.37 & 90.75 & 82.69 & 77.06 \\
FedPGP [ICML2024] & 89.04 & \thirdcell{95.10} & \thirdcell{96.96} & \thirdcell{93.70} & \secondcell{88.55} & \thirdcell{77.20} & \thirdcell{95.06} & \firstcell{93.77} & \thirdcell{88.65} & \secondcell{89.79} & \secondcell{77.68} & \thirdcell{87.91} & \thirdcell{53.08} & \secondcell{93.92} & \thirdcell{86.50} & \thirdcell{81.48} \\
FedDDA [ICML2025] & \secondcell{89.32} & \secondcell{97.14} & \secondcell{98.23} & \secondcell{94.90} & \thirdcell{87.07} & \secondcell{78.67} & \secondcell{96.32} & 93.52 & \secondcell{88.89} & \thirdcell{89.74} & \thirdcell{77.34} & \firstcell{88.46} & \firstcell{63.67} & \firstcell{94.14} & \secondcell{88.06} & \secondcell{83.57} \\
\midrule
\method (Ours) & \firstcell{90.57} & \firstcell{100.00} & \firstcell{100.00} & \firstcell{96.86} & \firstcell{88.60} & \firstcell{80.70} & \firstcell{97.40} & 93.40 & \firstcell{90.03} & \firstcell{90.30} & \firstcell{78.80} & \secondcell{88.40} & \secondcell{63.30} & \thirdcell{93.80} & \firstcell{88.10} & \firstcell{83.78} \\
\midrule
\multicolumn{17}{l}{\textit{Setting \textcircled{2}: one domain for two clients}} \\
Zero-shot CLIP [ICML2021] & 81.14 & 72.66 & 73.99 & 75.93 & 84.33 & 66.28 & 89.07 & 89.69 & 82.34 & 71.94 & 53.31 & 65.73 & 13.57 & 83.49 & 66.46 & 59.08 \\
PromptFL [TMC2023] & \secondcell{89.89} & 84.89 & 92.19 & 88.99 & 86.65 & 75.38 & 94.51 & \secondcell{93.49} & 87.51 & 86.57 & 71.29 & 81.90 & 35.34 & 92.40 & 80.30 & 74.63 \\
PromptFL+Prox [TMC2023; MLSys2020] & \thirdcell{89.25} & 86.85 & 91.80 & 89.30 & 86.13 & 74.60 & 94.40 & \thirdcell{93.44} & 87.15 & 86.48 & 72.63 & 81.72 & 34.43 & 92.31 & 80.85 & 74.73 \\
FedOTP [CVPR2024] & 84.73 & 89.16 & \thirdcell{95.94} & 89.94 & 77.96 & 73.96 & 91.20 & 87.10 & 82.56 & 85.73 & 69.50 & 81.45 & 48.49 & 90.38 & 82.11 & 76.27 \\
FedPGP [ICML2024] & 89.24 & \thirdcell{91.66} & 95.18 & \thirdcell{92.03} & \secondcell{87.92} & \thirdcell{77.01} & \thirdcell{94.96} & 92.71 & \thirdcell{88.15} & \secondcell{88.89} & \secondcell{77.37} & \secondcell{87.08} & \thirdcell{51.14} & \firstcell{93.70} & \secondcell{86.07} & \thirdcell{80.71} \\
FedDDA [ICML2025] & 88.39 & \secondcell{95.06} & \firstcell{98.10} & \secondcell{93.85} & \thirdcell{86.69} & \firstcell{77.98} & \firstcell{95.27} & 92.96 & \secondcell{88.22} & \thirdcell{88.65} & \thirdcell{76.07} & \thirdcell{86.67} & \firstcell{61.76} & \thirdcell{93.47} & \firstcell{86.80} & \secondcell{82.24} \\
\midrule
\method (Ours) & \firstcell{90.00} & \firstcell{98.10} & \secondcell{96.90} & \firstcell{95.00} & \firstcell{88.25} & \secondcell{77.05} & \secondcell{95.05} & \firstcell{93.60} & \firstcell{88.49} & \firstcell{89.65} & \firstcell{78.30} & \firstcell{87.15} & \secondcell{59.45} & \secondcell{93.65} & \thirdcell{85.95} & \firstcell{82.36} \\
\bottomrule
\end{tabular}%
}
\caption{Top-1 accuracy (\%) under domain shift. Setting~\textcircled{1} uses one client per domain and Setting~\textcircled{2} uses two. Brackets show publication venues. Pale red, yellow, and green mark the first-, second-, and third-best results; bold and underline additionally identify the top two.}
\pdfanchor{tab:domain-shift}\label{tab:domain-shift}
\end{table*}

\begin{table*}[!t]
\centering
\setlength{\tabcolsep}{5.2pt}
\renewcommand{\arraystretch}{1.12}
\resizebox{0.9\textwidth}{!}{%
\begin{tabular}{lccc|ccc|ccc}
\toprule
\multirow[c]{2}{*}{Method} & \multicolumn{3}{c}{Office31} & \multicolumn{3}{c}{OfficeHome} & \multicolumn{3}{c}{DomainNet100} \\
& $\beta=0.1$ & $\beta=0.3$ & $\beta=0.5$ & $\beta=0.1$ & $\beta=0.3$ & $\beta=0.5$ & $\beta=0.1$ & $\beta=0.3$ & $\beta=0.5$ \\
\midrule
Zero-shot CLIP [ICML2021] & 75.36 & 75.01 & 75.49 & 82.27 & 82.39 & 82.30 & 59.07 & 59.24 & 59.11 \\
PromptFL [TMC2023] & 89.01 & 90.44 & 88.40 & 87.35 & 87.23 & 87.01 & 73.14 & 73.88 & 74.23 \\
PromptFL+Prox [TMC2023; MLSys2020] & 89.27 & 89.66 & 88.11 & 87.46 & 87.32 & 87.36 & 73.41 & 73.66 & 74.07 \\
FedOTP [CVPR2024] & 90.78 & 90.54 & 88.75 & 84.81 & 85.66 & 84.28 & 77.52 & 77.70 & 76.94 \\
FedPGP [ICML2024] & \thirdcell{91.78} & \thirdcell{90.88} & \thirdcell{91.68} & \thirdcell{89.49} & \thirdcell{89.63} & \thirdcell{88.78} & \thirdcell{80.72} & \thirdcell{82.46} & \thirdcell{82.12} \\
FedDDA [ICML2025] & \secondcell{94.68} & \secondcell{94.34} & \secondcell{94.73} & \secondcell{89.85} & \secondcell{90.25} & \secondcell{89.24} & \secondcell{82.98} & \secondcell{83.24} & \secondcell{82.82} \\
\midrule
\method (Ours) & \firstcell{96.46} & \firstcell{96.02} & \firstcell{95.57} & \firstcell{90.65} & \firstcell{91.31} & \firstcell{90.05} & \firstcell{87.36} & \firstcell{85.99} & \firstcell{85.28} \\
\bottomrule
\end{tabular}%
}
\caption{Macro-client top-1 accuracy (\%) under joint domain and label shifts in Setting~\textcircled{2}. Brackets show publication venues. Pale red, yellow, and green mark the first-, second-, and third-best results; bold and underline additionally identify the top two.}
\pdfanchor{tab:label-shift}\label{tab:label-shift}
\vspace{-0.4em}
\end{table*}
\vspace{-0.43em}

\section{Experiments}

\subsection{Experimental Setup}
\paragraph{Datasets.}
We evaluate on three multi-domain classification benchmarks. Office31~\citep{saenko2010office31} contains 31 office-object categories from Amazon (A), Webcam (W), and digital single-lens reflex (DSLR; D). OfficeHome~\citep{venkateswara2017officehome} contains 65 categories from Art (A), Clipart (C), Product (P), and Real World (R). DomainNet~\citep{peng2019domainnet} contains six domains---Clipart (C), Infograph (I), Painting (P), Quickdraw (Q), Real (R), and Sketch (S). 

\vspace{-0.43em}
\paragraph{Data heterogeneity.}

We use two client partitions. Setting~\textcircled{1} assigns each domain and its full class distribution to one client. Setting~\textcircled{2} splits each domain across two clients, either evenly ($\beta=0$) or class-wise under a Dirichlet distribution with $\beta\in\{0.1,0.3,0.5\}$. Lower $\beta$ produces stronger label imbalance; full partition details are provided in the supplementary material.

\vspace{-0.45em}
\paragraph{Implementation details.}

All methods use a frozen CLIP ViT-B/16 backbone with 16 prompt tokens, SGD at learning rate $0.001$, batch size 32, and one local epoch per round. We report macro-client top-1 accuracy, defined as the unweighted mean of client-level test accuracies. We average results over three independent seeds. For each seed, the selected round maximizes macro-client test accuracy, and all domain-wise values come from that round. Further training details are provided in the supplementary material.

\vspace{-0.45em}
\paragraph{Comparison methods.}
We compare against Zero-shot CLIP~\citep{radford2021clip}, PromptFL~\citep{guo2023promptfl}, PromptFL+Prox~\citep{guo2023promptfl,li2020fedprox}, FedOTP~\citep{li2024fedotp}, FedPGP~\citep{cui2024fedpgp}, and \feddda~\citep{yang2025feddda}.

\begin{figure*}[t]
    \centering
    \includegraphics[width=0.78\textwidth]{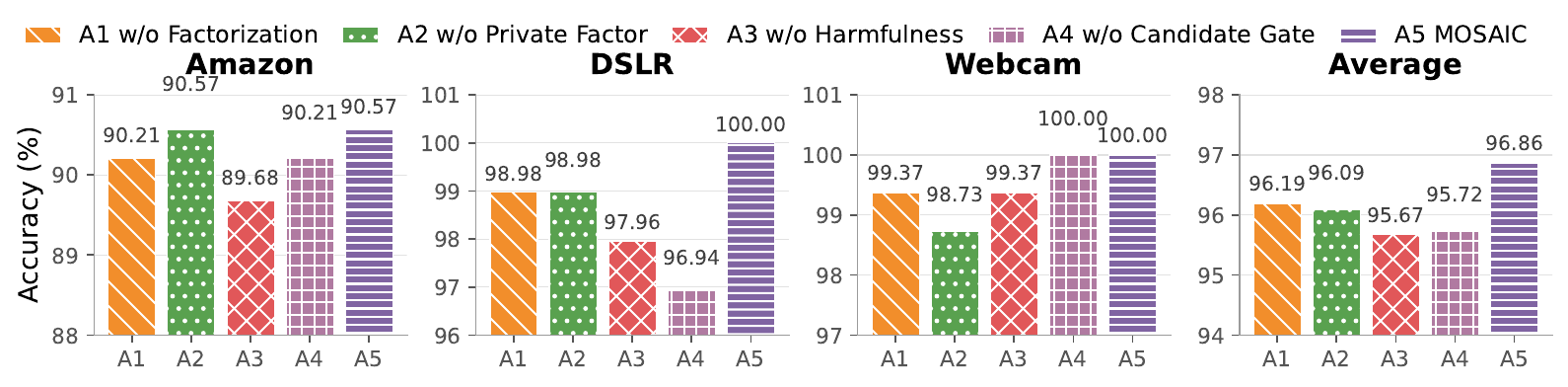}
    \caption{Component ablation on Office31 under Setting~\textcircled{1}. Each panel uses its own vertical scale to make the domain-wise differences visible. A5 denotes the complete \method configuration.}
    \pdfanchor{fig:ablation-office31}\label{fig:ablation-office31}
\end{figure*}

\begin{figure*}[t]
    \centering
    \includegraphics[width=0.9\textwidth]{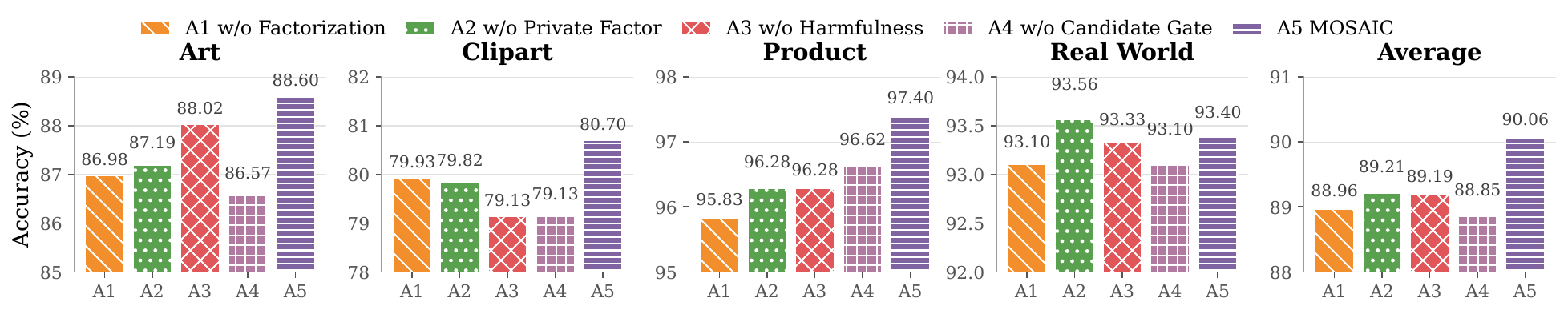}
    \caption{Component ablation on OfficeHome under Setting~\textcircled{1}. The complete configuration (A5) gives the highest macro-client average.}
    \pdfanchor{fig:ablation-officehome}\label{fig:ablation-officehome}
\end{figure*}

\begin{table}[h!]
\centering
\small
\setlength{\tabcolsep}{7pt}
\renewcommand{\arraystretch}{1.14}
\begin{tabular}{llcc}
\toprule
Axis & Value & Final & Best \\
\midrule
\multirow[c]{3}{*}{Rank $r$} & 8 & \underline{95.85} & 96.44 \\
         & 16 & 95.67 & \underline{96.52} \\
         & 32 & \textbf{96.43} & \textbf{96.86} \\
\midrule
\multirow[c]{3}{*}{Scale $\alpha$} & 0.03 & \underline{95.97} & \underline{96.69} \\
               & 0.05 & \textbf{96.50} & \textbf{96.80} \\
               & 0.10 & 95.67 & 96.52 \\
\midrule
\multirow[c]{3}{*}{Weight $\gamma$} & 1.0 & \textbf{95.67} & \textbf{96.52} \\
                 & 1.5 & \underline{95.58} & 95.79 \\
                 & 2.0 & 94.43 & \underline{95.91} \\
\midrule
\multirow[c]{4}{*}{Quantile $q$ ($\gamma=1.5$)} & 10\% & \textbf{96.10} & \textbf{96.74} \\
                              & 20\% & 95.58 & 95.79 \\
                              & 25\% & \underline{95.72} & 96.41 \\
                              & 50\% & 95.33 & \underline{96.60} \\
\bottomrule
\end{tabular}
\caption{Parameter sensitivity on Office31 Setting~\textcircled{1}. Bold and underline mark the best and second-best results.}
\pdfanchor{tab:parameter-sensitivity}\label{tab:parameter-sensitivity}
\end{table}

\subsection{Comparisons under Domain and Label Shifts}
\noindent\textbf{Superior macro-client accuracy across client-partition protocols.}
\tabref{tab:domain-shift} reports consistent macro-client gains for \method across all three datasets and both client-partition protocols. The largest gains occur on Office31, while the smaller but positive gains on OfficeHome and DomainNet100 show that the correction remains effective under more diverse visual shifts. The same ordering holds when two clients represent each domain, indicating that the shared/private decomposition is useful beyond the single-client-per-domain setting.

\noindent\textbf{Domain-wise gains with a balanced personalization trade-off.}
The domain-level results show a balanced personalization trade-off rather than uniform dominance. On OfficeHome Real World under Setting~\textcircled{1}, \method falls outside the top three at 93.40\%, trailing the best result by only 0.37 points, while its gains on the other three domains produce the highest OfficeHome macro-client average. \method improves all Office31 domains and most OfficeHome domains, while its DomainNet100 gains are concentrated in Clipart and Infograph. It does not win every domain, particularly on the most diverse benchmark; nevertheless, the macro-client improvement is consistent with selectively correcting decision-relevant shifts.

\noindent\textbf{Consistent accuracy under coupled domain and label shifts.}
\tabref{tab:label-shift} evaluates the multi-client-per-domain protocol under Dirichlet label imbalance. \method outperforms \feddda for every reported $\beta$ and dataset. The largest gains occur on DomainNet100, indicating that the factorized residual remains useful when broad visual diversity is coupled with client-level label imbalance.

\subsection{Component Ablation and Parameter Sensitivity}

\noindent\textbf{Complementary component benefits under heterogeneous domains.}\ 
Figures~\pdfjump{fig:ablation-office31}{\ref{fig:ablation-office31}} and~\pdfjump{fig:ablation-officehome}{\ref{fig:ablation-officehome}} isolate the proposed \mbox{components} in Setting~\textcircled{1}. A1 removes factorization, A2 removes the private domain factor, A3 removes harmful-pair emphasis, A4 replaces the candidate-wise gate with a fixed correction, and A5 is the complete model. A5 achieves the best macro-client average on both datasets; every removal degrades performance. The domain-wise bars show that these degradations vary by domain, supporting the intended roles of shared class structure, private domain activation, selective emphasis, and image-conditioned correction.

\noindent\textbf{Robust parameter sensitivity under targeted correction.}
\tabref{tab:parameter-sensitivity} shows moderate residual rank and correction scale are most effective. Aggressive intervention, through a large harmful-pair weight or broad pair selection, reduces accuracy. These trends support the intended design: correction should be expressive enough to model class-domain interactions, but applied selectively and with controlled strength.

\section{Conclusion}

We introduced \method for personalized federated VLM tuning under class-conditioned domain drift. \method separates the common domain effect from decision-harmful class-domain residuals and selectively corrects them using shared class structure and private domain activation. The main results and component ablations consistently support this targeted correction.
\noindent\textbf{Impact and Future Work.}
By correcting harmful interactions without sharing raw images, \method supports federated VLM adaptation across heterogeneous institutions. Future work will scale harmful-set construction to larger or unseen domains and protect exchanged statistics through secure aggregation or differential privacy. We will extend this perspective to prototype learning by constructing dedicated benchmarks with explicitly curated harmful sets, enabling systematic evaluation of how such interactions affect existing prototype-based methods.

\bibliography{ref}

% Supplementary material (the complete standalone supplement body).
\clearpage
\setcounter{section}{0}
\setcounter{secnumdepth}{2}
\begin{center}
{\LARGE\bfseries Supplementary Material\par}
\end{center}

\appendix
\section{Guide to the Supplementary Material}

This supplementary material supports the main paper through a unified line of evidence: the motivating phenomenon is genuine and general, the harmfulness score identifies decision-relevant drift more effectively than residual magnitude, and the resulting method is lightweight, evaluated under a consistent protocol, and reproducible. Section~B details the datasets, client-partition protocols, handling of missing class--domain pairs, optimization settings, and dataset-specific hyperparameters. Section~C provides detailed descriptions of the comparison methods and contrasts the granularity of their correction mechanisms. Section~D analyzes the parameter, communication, and computational costs of MOSAIC, together with the privacy implications of the exchanged statistics. Section~E extends the qualitative evidence of class-conditioned domain drift to all three benchmarks. Section~F presents the training-only diagnostic analysis, including residual construction, the decision-aware harmfulness score and its counterfactual audit, robustness to the choice of diagnostic representation, and harmfulness-quantile trends and representative cases that motivate the two-level selective-correction design. Finally, Section~G reports complete domain- and class-level results, discusses domain-specific regressions, and clarifies the scope of the empirical claims.
\section{Detailed Experimental Setup}
\label{sec:supp-experimental-setup}

\paragraph{Datasets.}
We evaluate on Office31~\citep{saenko2010office31}, OfficeHome~\citep{venkateswara2017officehome}, and DomainNet~\citep{peng2019domainnet}. Office31 contains 31 categories from Amazon, Webcam, and DSLR; OfficeHome contains 65 categories from Art, Clipart, Product, and Real World; and DomainNet contains Clipart, Infograph, Painting, Quickdraw, Real, and Sketch, each with 345 categories. To enhance evaluation efficiency, we follow the evaluation protocol of \citet{yang2025feddda} and use the first 100 categories of every domain, denoted DomainNet100; \method and all compared methods are evaluated on this identical subset.

\paragraph{Client partitions.}
The two protocols separate cross-domain heterogeneity from variation among clients sharing a visual domain. In the single-client-per-domain protocol (Setting~\textcircled{1}), each domain is assigned to one client, which retains the complete class distribution of that domain. In the multi-client-per-domain protocol (Setting~\textcircled{2}), every domain is split class-wise across two clients. For the domain-only study, samples are divided evenly ($\beta=0$), so the two clients share a domain style but observe disjoint samples. For the coupled domain--label-shift study, the samples of each class are allocated with a Dirichlet distribution using $\beta\in\{0.1,0.3,0.5\}$; a smaller $\beta$ produces stronger client-level label imbalance while preserving the visual-domain assignment.

Dirichlet allocation does not require every client to contain every class. For statistic construction, a client computes a training-feature sum and count only for its locally observed classes. Clients belonging to the same visual domain are combined by summing these sufficient statistics, so the resulting class--domain mean is the sample-count-weighted mean across those clients. A client with zero examples of class $c$ contributes neither a feature sum nor a count for $(c,d)$. If the pooled count of a class--domain pair is zero, that pair is omitted from residual and harmful-set construction; it is never represented by a zero feature vector. The reported datasets retain nonzero pooled training counts for all evaluated pairs, so no low-count shrinkage or test-set imputation is used. More generally, a deployment may omit pairs below a prespecified training-count threshold or shrink their means toward the domain mean, but must not use held-out test images to fill missing classes.

\paragraph{Optimization and reproducibility.}

All methods use CLIP with a frozen ViT-B/16 image encoder. We use a prompt length of 16, stochastic gradient descent with learning rate $0.001$, batch size 32, and one local epoch per communication round. Following the official FedDDA implementation~\citep{yang2025feddda}, we use its original training/test splits and client-partition protocol, and no separate validation split is constructed. For each client, ordinary top-1 accuracy is computed over all of its test samples. Macro-client accuracy is the unweighted arithmetic mean of these client-level accuracies; because every domain contains the same number of clients, it is also equal to the unweighted mean of domain-level accuracies. It is neither class-balanced accuracy nor pooled sample-weighted accuracy. The main comparison and parameter-sensitivity results average three independent random seeds. Before each run, the implementation seeds Python, NumPy, PyTorch, and all CUDA generators. As in the official implementation, macro-client test accuracy is evaluated after every communication round and each run reports its maximum. This reporting rule is inherited from the compared protocol and is applied identically to \method and every compared method, so all numbers in one table are produced under the same protocol; it determines only which round is reported and never influences residual estimation, harmfulness scoring, or harmful-set construction, which use training data exclusively (Sec.~\ref{sec:supp-diagnosis}). Tables report the mean of these seed-level results; domain-wise values average clients assigned to the same domain at each seed-specific selected round before averaging across seeds. \tabref{tab:supp-dataset-hyperparameters} reports the residual-module hyperparameters used for each dataset and the candidate values considered in the completed staged searches.

\noindent\textbf{Compute resources.}
All experiments were conducted on a GPU server with two custom-modified NVIDIA RTX 4090 GPUs, each upgraded to 48\,GB memory. The server has a 24-core QEMU virtual CPU and 62\,GB system memory. The software environment uses Ubuntu 22.04 with CUDA 12.1, cuDNN 8, Python 3.12, and PyTorch 2.3.0. Each experimental run uses simulated federated clients on the same server.

\begin{table*}[t]
\centering
\small
\setlength{\tabcolsep}{7pt}
\begin{tabular}{lccccc}
\toprule
Dataset & Rank $r$ & Scale $\alpha$ & Harmful weight $\gamma$ & Quantile $q$ & Regularization $\lambda$ \\
\midrule
Office31 & 32 & 0.10 & 1.5 & 20\% & $10^{-4}$ \\
OfficeHome & 16 & 0.05 & 1.5 & 20\% & $10^{-4}$ \\
DomainNet100 & 32 & 0.10 & 1.5 & 20\% & $10^{-4}$ \\
\midrule
Candidate values & $\{8,16,32\}$ & $\{0.03,0.05,0.10\}$ &
$\{1.0,1.5,2.0\}$ & $\{10,20,25,50\}\%$ &
$\{10^{-5},10^{-4},5{\times}10^{-4}\}$ \\
\bottomrule
\end{tabular}
\caption{Per-dataset residual-module hyperparameters. The final rows give the configurations used in the reported experiments. The candidate row lists values considered across the completed staged searches; it is not an exhaustive Cartesian-product sweep on every dataset.}
\pdfanchor{tab:supp-dataset-hyperparameters}\label{tab:supp-dataset-hyperparameters}
\end{table*}

\section{Detailed Comparison Baselines}
\label{sec:supp-baselines}

This section expands the comparison-method summary in the main paper. The selected baselines cover an unadapted vision-language model (VLM), federated prompt learning, optimization-level stabilization, global--local prompt interaction, prompt personalization, and joint text--visual personalization. All trainable methods use the same frozen CLIP backbone and the same client partitions as \method; consequently, the comparison isolates the adaptation and personalization mechanisms rather than differences in pretrained representations.

\paragraph{Zero-shot CLIP.}
Zero-shot CLIP~\citep{radford2021clip} forms one textual prototype per class from a hand-crafted prompt and predicts by image--text similarity. It performs no downstream or federated training and therefore provides a reference for the transferable knowledge already present in the pretrained model. Because neither the text representation nor the image representation adapts to client data, it cannot accommodate local acquisition styles or class-conditioned domain effects.

\paragraph{PromptFL.}
PromptFL~\citep{guo2023promptfl} converts prompt learning into a federated parameter-efficient optimization problem. Each client updates a short sequence of continuous textual context tokens while the CLIP encoders remain frozen; the server averages these prompt parameters using a FedAvg-style rule. Communication is compact because clients exchange prompts instead of full VLM weights. The resulting prompt is nevertheless global: domain-specific evidence is mixed into a single textual adaptation state, without a private visual correction for an individual client or class-domain pair.

\paragraph{PromptFL+Prox.}
PromptFL+Prox augments PromptFL with the proximal regularizer of FedProx~\citep{li2020fedprox}. During local training, the client objective penalizes the distance between the local prompt and the current global prompt. The penalty limits client drift under heterogeneous data and changes the optimization dynamics, but not the representation granularity: a single global prompt still absorbs all domain effects. This baseline tests whether improved local stability alone explains the gains of \method.

\paragraph{FedOTP.}
FedOTP~\citep{li2024fedotp} maintains global and local prompts and uses unbalanced optimal transport to model their interaction. The global prompt transfers shared knowledge across clients, whereas the local prompt captures client-specific information; transport-based alignment allows the two prompt sets to cooperate without forcing a rigid one-to-one correspondence. Personalization is therefore more expressive than global prompt averaging, but it remains prompt-side and client-level. It does not explicitly estimate a visual residual for class $c$ in domain $d$ or test how that residual changes the true-versus-competitor margin.

\paragraph{FedPGP.}
FedPGP~\citep{cui2024fedpgp} develops guided prompt personalization for federated VLMs. It preserves global CLIP knowledge while learning a compact personalized prompt component, using low-rank structure to limit the local trainable state and reduce overfitting. Its central question is how to balance shared and personalized prompt knowledge. In contrast, \method asks which class-domain visual interaction is decision-harmful and corrects the candidate feature only when activated by an image-conditioned gate.

\paragraph{FedDDA.}
FedDDA~\citep{yang2025feddda} is the closest baseline. It decouples globally shared and client-private textual priors and combines shared and specific visual adapters through a dynamic gate. This design directly addresses personalized federated VLM tuning under domain shift and provides the strong shared/private foundation used in our comparison. Its specialization unit is still the client domain: the visual mixture is shared across the semantic candidates of an image. \method adds the missing class-domain interaction by factorizing a candidate-specific residual into globally shared class structure, a client-private domain factor, and a shared basis. A decision-aware score and candidate-wise gate then determine where the additional correction is warranted.

\paragraph{Mechanistic coverage of the comparison methods.}
\tabref{tab:supp-baseline-mechanisms} consolidates these distinctions. The progression from global prompt averaging to client-private adaptation improves personalization, but only \method combines private domain state with an explicitly decision-aware, image--candidate correction. The table clarifies that the comparison is not simply between parameter counts: the methods differ in where adaptation occurs and in the granularity at which domain-dependent errors can be corrected.

\begin{table*}[t]
\centering
\small
\setlength{\tabcolsep}{4pt}
\begin{tabular}{m{0.12\textwidth}m{0.13\textwidth}m{0.23\textwidth}m{0.22\textwidth}m{0.15\textwidth}}
\toprule
Method & Adapted state & Shared mechanism & Private mechanism & Correction granularity \\
\midrule
Zero-shot CLIP & None & Pretrained image--text space & None & None \\
PromptFL & Text prompt & Federated prompt averaging & None & Task/global prompt \\
PromptFL+Prox & Text prompt & Prompt averaging with proximal control & None & Task/global prompt \\
FedOTP & Text prompts & Global prompt & Local prompt with optimal transport & Client/prompt \\
FedPGP & Text prompts & CLIP-guided global knowledge & Low-rank personalized prompt & Client/prompt \\
FedDDA & Text and visual adapters & Global prompt and shared adapter & Local prompt and specific adapter & Client/domain \\
\method & Text, visual, and residual adapters & Class factor and residual basis & Domain factor and local adapters & Image--candidate pair \\
\bottomrule
\end{tabular}
\caption{Mechanistic scope of the comparison methods. ``Private'' denotes a client-retained adaptation state.}
\pdfanchor{tab:supp-baseline-mechanisms}\label{tab:supp-baseline-mechanisms}
\end{table*}

\section{Parameter, Communication, and Computational Cost}
\label{sec:supp-efficiency}

Let $C$ be the number of classes, $D=512$ the CLIP feature dimension, and $r$ the residual rank. The factorized correction adds $Cr$ class-factor parameters, two $Dr$ matrices for the residual basis and image projection, and $C$ class amplitudes. Its globally shared parameter count is therefore
\begin{equation}
P_{\mathrm{res}}=Cr+2Dr+C,
\label{eq:supp-residual-params}
\end{equation}
while each domain conceptually retains one private $r$-dimensional factor. These counts exclude the frozen CLIP backbone and the prompts/adapters already present in the shared--private VLM adaptation model.

\begin{table}[t]
\centering
\small
\setlength{\tabcolsep}{2.5pt}
\begin{tabular}{lrrrr}
\toprule
Dataset & $(C,r)$ & Shared params & Round trip & Statistics \\
 & & & (KiB) & (KiB) \\
\midrule
Office31 & $(31,32)$ & 33,791 & 264.0 & 62.1 \\
OfficeHome & $(65,16)$ & 17,489 & 136.6 & 130.3 \\
DomainNet100 & $(100,32)$ & 36,068 & 281.8 & 200.4 \\
\bottomrule
\end{tabular}
\caption{Residual-module storage and communication overhead. Communication assumes FP32 and reports one client upload plus one server download per round. Statistics are a one-time per-domain upper bound when all classes are observed.}
\pdfanchor{tab:supp-efficiency}\label{tab:supp-efficiency}
\end{table}

The domain factor is never uploaded. Per-round residual communication consists only of the parameters in Eq.~\ref{eq:supp-residual-params}; the values in \tabref{tab:supp-efficiency} count both upload and download and are therefore $8P_{\mathrm{res}}$ bytes. Harmful-set construction requires a one-time exchange of class-feature sums and counts. If $K_d$ classes are observed in domain $d$, its payload is approximately $4K_d(D+1)$ bytes under FP32 storage; the table reports the upper bound $K_d=C$. Raw images are never included in this payload. Secure aggregation is compatible with sums and counts, although the present experiments do not implement secure aggregation and therefore do not claim that these statistics are free of privacy risk.

For a batch of size $B$, the current vectorized implementation forms candidate residuals with cost $\mathcal O(BCDr)$ and then computes corrected candidate features and logits with cost $\mathcal O(BCD)$. The cost is consequently linear in the class count, which is why we explicitly evaluate $C=31$, $65$, and $100$. Because the residual direction for a fixed class and domain is image-independent before gating, an optimized implementation can cache the $C\times D$ projected residuals for each domain. This reduces the repeated projection to $\mathcal O(CDr)$ per cache update and leaves an online cost of $\mathcal O(B(Dr+Cr+CD))$. The candidate dimension is vectorized rather than evaluated with $C$ separate encoder passes, and the frozen image encoder remains the dominant end-to-end component.

\section{Additional Qualitative Evidence of Class-Conditioned Domain Drift}
\label{sec:supp-qualitative}

\begin{figure*}[t]
    \centering
    \begin{minipage}[t]{0.41\textwidth}
        \centering
        \includegraphics[width=\linewidth]{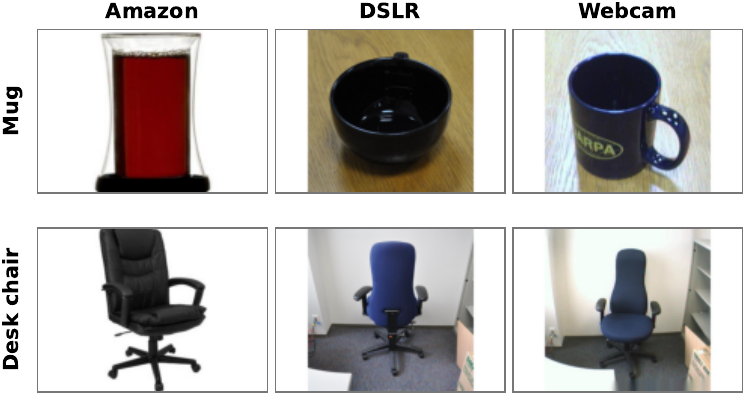}\\[-0.2em]
        \textbf{(a) Office31}
    \end{minipage}\hfill
    \begin{minipage}[t]{0.56\textwidth}
        \centering
        \includegraphics[width=\linewidth]{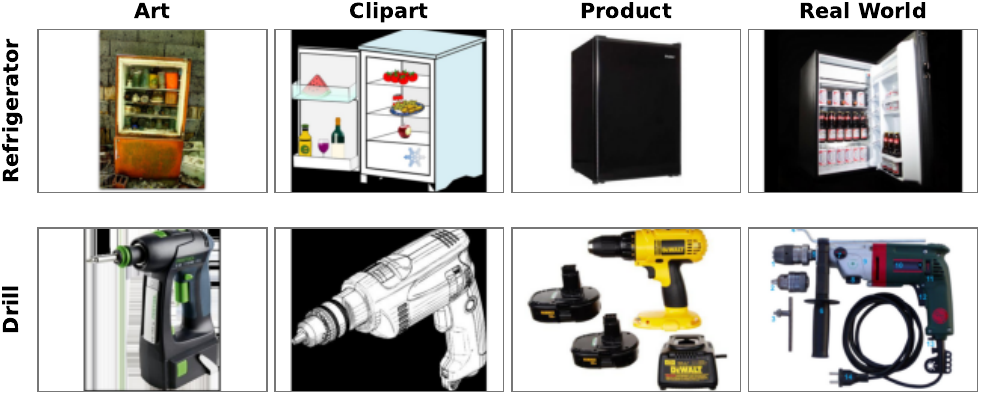}\\[-0.2em]
        \textbf{(b) OfficeHome}
    \end{minipage}
    \vspace{0.5em}

    \includegraphics[width=0.98\textwidth]{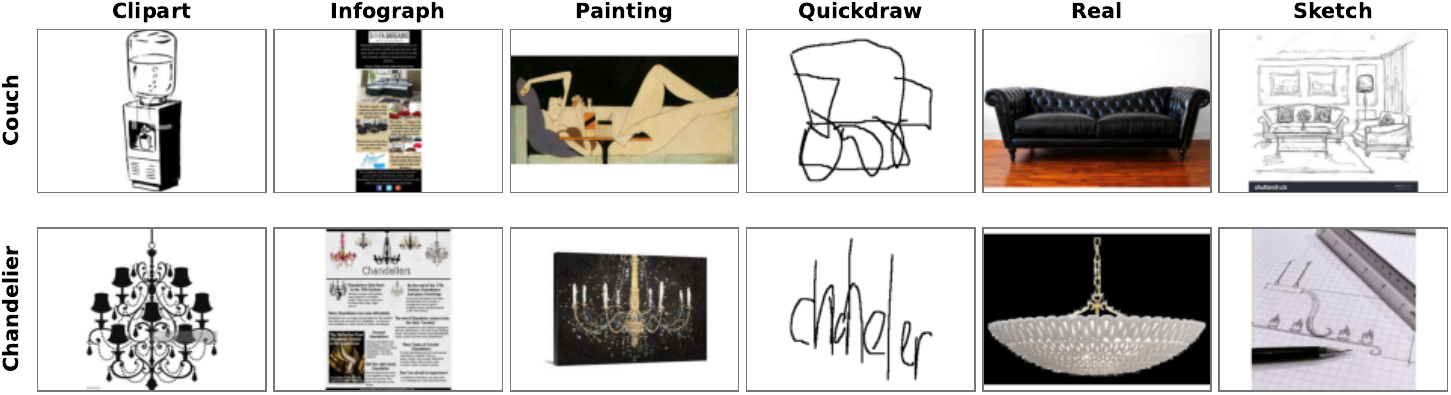}\\[-0.2em]
    \textbf{(c) DomainNet100}
    \caption{Additional qualitative evidence of class-conditioned domain drift. Classes form rows and domains form columns. The same domain changes different semantic classes through different visual cues, rather than through one class-agnostic offset.}
    \pdfanchor{fig:supp-all-dataset-examples}\label{fig:supp-all-dataset-examples}
\end{figure*}

The main paper visualizes class-conditioned appearance changes on DomainNet100. Here we extend that observation to every evaluated benchmark. For each dataset, we select two classes present in every domain and display one deterministic representative image from each class-domain pair. The representative-image rule uses only image contrast, edge strength, and non-blank area; it does not access predictions, test errors, or harmfulness scores. These examples are therefore intended as qualitative evidence of non-additive appearance changes rather than as an accuracy comparison. \figref{fig:supp-all-dataset-examples} presents the selected pairs across all three benchmarks.

\figref{fig:supp-all-dataset-examples} arranges the three datasets in a compact two-column layout. In Office31, Amazon largely removes scene context, whereas DSLR and Webcam introduce desk surfaces, viewpoint changes, and camera-dependent image quality; the mug shift is dominated by material and local background, while the chair shift changes pose and geometry. In OfficeHome, Clipart simplifies refrigerators into large planar outlines but retains fine mechanical contours for drills. DomainNet100 further shows that Quickdraw, Sketch, Infograph, and photographic domains preserve different cues for couches and chandeliers. The phenomenon motivating \method is therefore present in all three benchmarks and is not restricted to the dog/chair example in the main paper.

\section{Training-Only Diagnosis of Harmful Class-Domain Residuals}
\label{sec:supp-diagnosis}

This section validates whether the observed class-dependent changes are geometrically measurable and decision-relevant under a strict train/test separation for all diagnostic quantities: residuals, harmfulness scores, and the harmful set are estimated exclusively from deterministic training subsets, while the held-out test split is used only to measure the resulting prediction outcomes and counterfactual changes. The test split never enters residual estimation, scoring, or harmful-set construction.

\subsection{Leave-One-Domain-Out Residual}

Let $\mu_{c,d}$ be the mean normalized image feature of class $c$ in domain $d$, computed from training images. We first compare this mean with the same class in all other domains:
\begin{equation}
\delta_{c,d}=\mu_{c,d}-\mu_{c,-d}, \qquad
\mu_{c,-d}=\frac{1}{|\mathcal D|-1}\sum_{d'\ne d}\mu_{c,d'}.
\label{eq:supp-loo-shift}
\end{equation}
The displacement $\delta_{c,d}$ contains both a common domain effect and the interaction specific to class $c$. We remove the average displacement of all other classes in the same domain:
\begin{equation}
\widehat r_{c,d}=\delta_{c,d}-\frac{1}{C-1}\sum_{c'\ne c}\delta_{c',d}.
\label{eq:supp-residual}
\end{equation}
Equation~\ref{eq:supp-residual} is a leave-one-class-out estimate of the non-additive class-domain interaction. The analysis uses at most 80 deterministic training images per pair for Office31 and OfficeHome and at most 100 for DomainNet100. This produces 93, 260, and 600 class-domain pairs, respectively.

\subsection{Decision-Aware Score and Held-Out Counterfactual Audit}

\begin{figure*}[t]
    \centering
    \includegraphics[width=0.94\textwidth]{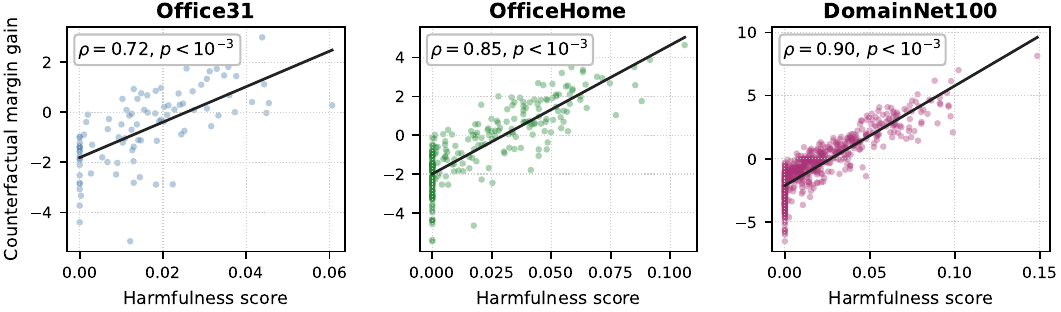}
    \caption{Pair-level harmfulness versus held-out counterfactual margin gain. Each point is one class-domain pair; the residual is computed on training data and evaluated on held-out test images. Spearman rank correlations are shown in each panel.}
    \pdfanchor{fig:supp-harmfulness-diagnostic}\label{fig:supp-harmfulness-diagnostic}
\end{figure*}

Residual magnitude alone does not indicate whether the displacement changes a classification decision. Given the local text prototypes $\{t_k\}_{k=1}^{C}$, we therefore compute
\begin{equation}
h(c,d)=\max\!\left(0,\max_{k\ne c}
\left[\langle t_k,\widehat r_{c,d}\rangle-
\langle t_c,\widehat r_{c,d}\rangle\right]\right).
\label{eq:supp-harmful-score}
\end{equation}
A positive value means that the residual favors at least one competitor over the true class. To audit this interpretation on held-out images, we construct an oracle diagnostic that subtracts the training-derived true class-domain residual,
\begin{equation}
v_x^{\mathrm{cf}}=\frac{v_x-\widehat r_{y,d}}{\|v_x-\widehat r_{y,d}\|_2},
\qquad
\Delta m=m(v_x^{\mathrm{cf}},y)-m(v_x,y),
\label{eq:supp-counterfactual}
\end{equation}
where $m(v,y)$ is the true-class logit minus the largest competing logit. This oracle uses the ground-truth class and is only a diagnostic; it is not an inference procedure and is not used to report \method accuracy. \figref{fig:supp-harmfulness-diagnostic} visualizes the pair-level relationship, and \tabref{tab:supp-diagnostic-correlations} provides the corresponding rank-correlation summary.

\paragraph{Scope of the oracle audit.}
Equation~\ref{eq:supp-counterfactual} asks whether removing a training-derived class--domain residual improves the held-out decision margin when the true class is known for analysis. This controlled intervention validates the direction targeted by the harmfulness score. During actual \method inference, the ground-truth label is unavailable and never enters the correction; the model independently constructs a correction and logit for every candidate class as described in Eqs.~(10)--(13) of the main paper.

\paragraph{Geometric consistency of the score.}
The harmfulness score and oracle margin change evaluate the same residual geometry from complementary views: $h(c,d)$ measures its projection along true-versus-competitor directions, while $\Delta m$ measures the resulting margin response after removal. Accordingly, this audit is a geometric consistency check for the scoring rule. Within this scope, \figref{fig:supp-harmfulness-diagnostic} and \tabref{tab:supp-diagnostic-correlations} show that $h(c,d)$ ranks the response to oracle removal more directly than residual norm: the rank correlation increases from $0.724$ on Office31 to $0.897$ on DomainNet100, whereas residual norm has a weaker and dataset-dependent relationship with the same outcome.

\paragraph{Interpretation boundary with respect to test error.}
The direct association with held-out error is weaker and less stable on small, low-error benchmarks: the Spearman correlations are $-0.144$, $0.364$, and $0.619$ on Office31, OfficeHome, and DomainNet100, respectively. This ordering follows the scale of the diagnostic rather than a failure of the scoring rule. Office31 contributes only 93 class--domain pairs over 31 classes and three visually close domains, and its adapted error rate is already low, so the held-out error of most pairs is compressed near zero; a rank correlation computed on such a low-variance, near-degenerate outcome is dominated by a handful of misclassified samples and can flip sign when a few predictions change. As the number of pairs, classes, and rendering styles grows---260 pairs over 65 classes in OfficeHome and 600 pairs over 100 classes in DomainNet100---the error outcome gains enough variance for the association to emerge, and it becomes positive and monotonically stronger. We therefore do not interpret $h(c,d)$ as a universal causal error predictor or assume that every high-score pair must have high test error, particularly on small benchmarks where the error signal itself is scarce. The score instead identifies a potentially adverse residual geometry; realized error also depends on sample count, within-domain variance, class difficulty, and the surrounding competitor structure. Moreover, subtracting every residual indiscriminately decreases the aggregate margin on all three datasets because many low-harm pairs have negative counterfactual gains. This observation motivates selective rather than wholesale correction.

\subsection{Construction and Fixed Use of the Harmful Set}

The harmful set is constructed once, offline, before \method optimization. We compute the required quantities from deterministic subsets of the training split, evaluate Eq.~\ref{eq:supp-harmful-score}, and select the top-$q$ class--domain pairs independently within each domain. This domain-balanced selection prevents a visually diverse domain from dominating a single global ranking. For the reported \method runs, we use the fixed round-49 FedDDA diagnostic space so that the class--domain means and local text prototypes match the shared/private adaptation foundation. The resulting set $\mathcal H$ remains fixed throughout federated training, avoiding online feedback and round-to-round selection drift. Held-out test labels are used only for the oracle audit and correlation analyses, never to construct $\mathcal H$.

\paragraph{Diagnostic-space robustness.}
To test whether the diagnosed geometry is created only by the selected FedDDA checkpoint, we repeat the complete audit using the frozen CLIP ViT-B/16 representation before federated adaptation. Frozen text prototypes average four standard prompt templates, and the residual statistics use at most 80 training images per class--domain pair on Office31 and OfficeHome and 100 on DomainNet100. \tabref{tab:supp-frozen-clip-robustness} shows that the frozen-space harmfulness score remains strongly associated with held-out counterfactual margin gain on all three datasets. The top-20\% sets also share 9, 27, and 69 pairs with their equal-size FedDDA counterparts. Thus, the adverse class-conditioned geometry is already measurable in frozen CLIP, while the moderate set overlap indicates that exact pair membership remains partly representation-dependent. We use the FedDDA-space set in the reported runs for alignment with the adapted shared/private feature space, rather than because the phenomenon is absent from the pretrained representation.

\begin{table}[t]
\centering
\scriptsize
\setlength{\tabcolsep}{2.5pt}
\begin{tabular}{lrrrr}
\toprule
Dataset & $\rho_{\mathrm{CLIP}}$ & $\rho_{\mathrm{FedDDA}}$ & Intersection/size & Jaccard \\
\midrule
Office31 & 0.656 & 0.724 & 9/18 & 0.333 \\
OfficeHome & 0.908 & 0.852 & 27/52 & 0.351 \\
DomainNet100 & 0.930 & 0.897 & 69/120 & 0.404 \\
\bottomrule
\end{tabular}
\caption{Robustness to the diagnostic representation. $\rho_{\mathrm{CLIP}}$ and $\rho_{\mathrm{FedDDA}}$ are Spearman correlations between $h$ and held-out $\Delta m$ in frozen-CLIP and round-49 FedDDA spaces. Set overlap compares domain-balanced top-20\% harmful sets; both sets have the size shown after the slash.}
\pdfanchor{tab:supp-frozen-clip-robustness}\label{tab:supp-frozen-clip-robustness}
\end{table}

\subsection{Harmfulness-Quantile Trends and Cases}

\begin{figure*}[t]
    \centering
    \includegraphics[width=0.92\textwidth]{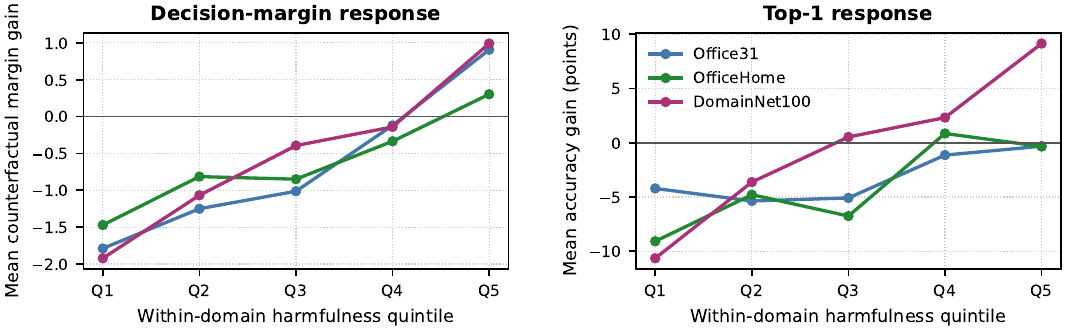}
    \caption{Domain-balanced harmfulness-quintile analysis. Values are unweighted pair means. Higher harmfulness generally corresponds to a more favorable response to oracle residual removal, while low-score pairs are frequently damaged by indiscriminate correction.}
    \pdfanchor{fig:supp-quantile-trends}\label{fig:supp-quantile-trends}
\end{figure*}

We divide the class-domain pairs into five harmfulness quintiles independently within each domain. This preserves the domain-balanced selection rule used by \method and prevents a visually diverse domain from dominating a global ranking. \figref{fig:supp-quantile-trends} shows a monotonic transition from harmful removal in the lower quintiles to useful removal in Q5. The transition is clearest on DomainNet100, where Q5 improves both the average margin and top-1 accuracy, while Office31 mainly exhibits a margin benefit. Thus, a high score identifies where correction is more promising, but it does not guarantee a top-1 flip for every pair.

\begin{figure*}[t]
    \centering
    \includegraphics[width=0.9\textwidth]{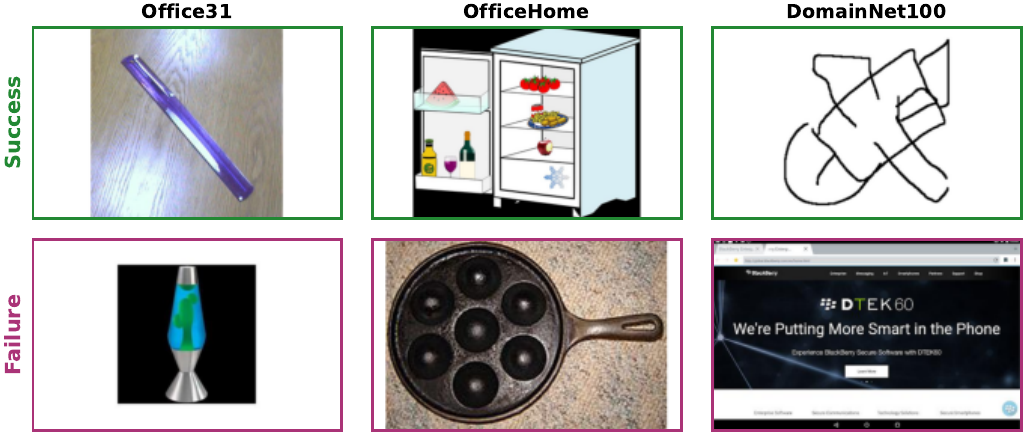}
    \caption{Representative Q5 class-domain pairs. Columns correspond to Office31, OfficeHome, and DomainNet100; the top and bottom rows show success and failure cases, respectively. Pair identities and metrics are given in the surrounding text. Displayed images are representative examples, not individually scored predictions.}
    \pdfanchor{fig:supp-quantile-cases}\label{fig:supp-quantile-cases}
\end{figure*}

To make the within-Q5 variation concrete, \figref{fig:supp-quantile-cases} shows one success and one failure case for each dataset. The panels retain only dataset names; the pair identities and quantitative values are reported below so that the image evidence remains legible.

\figref{fig:supp-quantile-cases} illustrates the remaining heterogeneity inside Q5. In Office31, the Webcam--ruler success pair has $h=0.044$, $\Delta m=+2.99$, and $\Delta\mathrm{Acc}=+0.0$ points, whereas the Amazon--desk-lamp failure pair has $h=0.045$, $\Delta m=-0.03$, and $\Delta\mathrm{Acc}=+5.3$ points. The latter shows that a small number of top-1 flips can coexist with a slightly negative pair-average margin response. In OfficeHome, Clipart--refrigerator responds positively ($h=0.106$, $\Delta m=+4.65$, $\Delta\mathrm{Acc}=+12.5$), while Real-World--pan is harmed by fixed removal ($h=0.007$, $\Delta m=-2.56$, $\Delta\mathrm{Acc}=+0.0$). The contrast is strongest on DomainNet100: Quickdraw--aircraft-carrier improves by $\Delta m=+8.15$ and $\Delta\mathrm{Acc}=+86.0$ points at $h=0.148$, whereas real--blackberry decreases by $\Delta m=-3.67$ and $\Delta\mathrm{Acc}=-17.5$ points at $h=0.011$. The latter also exposes semantic ambiguity because the DomainNet category blackberry includes brand-related imagery rather than only fruit appearance. Its presence in Q5 despite a near-zero score reflects the domain-balanced selection rule itself: within-domain quantile selection can admit near-zero-score pairs in domains whose score distribution is compressed toward zero. This is precisely the failure mode that the two-level design of \method is built to absorb---pair-level selection only proposes where correction may be needed, while the image-conditioned candidate gate and harmful-pair reweighting determine whether and how strongly correction acts on each image. These cases therefore explain why \method combines pair-level selection with an image-conditioned candidate gate instead of applying a constant correction to all images in a selected pair.

\begin{figure*}[t]
    \centering
    \includegraphics[width=0.9\textwidth]{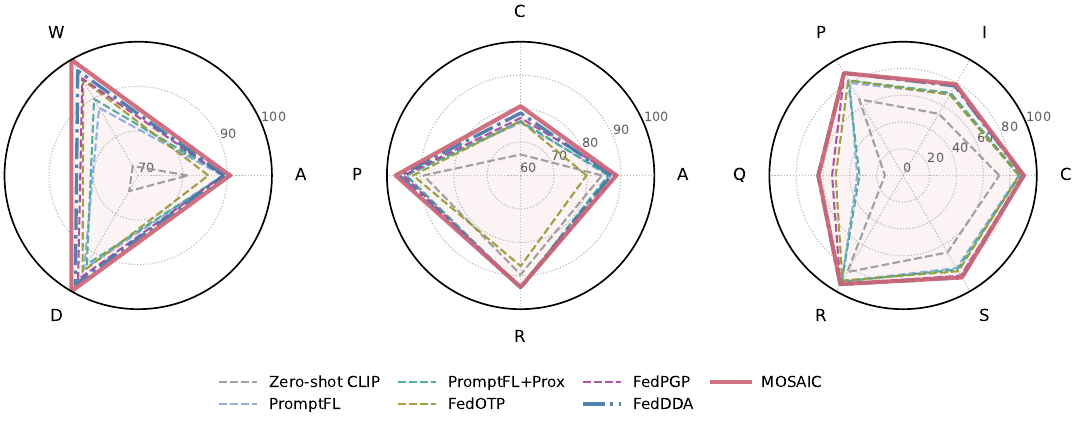}
    \caption{Domain-level accuracy profiles of all comparison methods under Setting 1, using the values reported in the main tables. From left to right, the three panels correspond to Office31, OfficeHome, and DomainNet100; domain abbreviations follow the main paper. MOSAIC and FedDDA are emphasized with solid and dash-dotted contours, respectively.}
    \pdfanchor{fig:supp-domain-radar}\label{fig:supp-domain-radar}
\end{figure*}

\section{Complete Domain- and Class-Level Results}
\label{sec:supp-granular-results}

The complete domain-level comparison in \tabref{tab:supp-domain-results} separates macro-client improvement from domain-specific variation. In Setting 1, MOSAIC improves every Office31 domain and raises the macro-client average by 1.96 points. OfficeHome gains are concentrated in Clipart (+2.03), Art (+1.53), and Product (+1.08), while Real World decreases by only 0.12 points. On DomainNet100, Clipart and Infograph improve by 0.56 and 1.46 points, respectively; the other changes are within 0.37 points. The macro-client improvement therefore comes from correcting selected domain-specific weaknesses rather than uniformly shifting every domain.

Setting 2 is more heterogeneous because two clients represent each visual domain. MOSAIC retains macro-client gains of 1.15, 0.27, and 0.12 points on Office31, OfficeHome, and DomainNet100. The largest positive changes occur on Office31 Webcam (+3.04) and DomainNet100 Infograph (+2.23). The method is not uniformly superior: DSLR decreases by 1.20 points, and DomainNet100 Quickdraw decreases by 2.31 points. These regressions are visible in \figref{fig:supp-domain-radar} and limit the empirical claim to a better macro-client trade-off rather than dominance on every client domain.

The class-level tables report actual \method predictions from saved terminal checkpoints at rounds 79, 59, and 79 for Office31, OfficeHome, and DomainNet100, respectively. These checkpoints provide a terminal-stability diagnostic and are distinct from the seed-specific rounds selected for the main comparison.

\begin{table}[t]
\centering
\small
\setlength{\tabcolsep}{4pt}
\begin{tabular}{lrrrr}
\toprule
Dataset & Pairs & $\rho(h,\Delta m)$ & $\rho(\|r\|,\Delta m)$ & $\rho(h,e)$ \\
\midrule
Office31    & 93  & 0.724 & $-0.275$ & $-0.144$ \\
OfficeHome  & 260 & 0.852 & 0.207 & 0.364 \\
DomainNet100 & 600 & 0.897 & 0.145 & 0.619 \\
\bottomrule
\end{tabular}
\caption{Pair-level Spearman correlations. $h$ is the harmfulness score, $\|r\|$ is residual norm, $e$ is held-out error, and $\Delta m$ is counterfactual margin gain.}
\pdfanchor{tab:supp-diagnostic-correlations}\label{tab:supp-diagnostic-correlations}
\end{table}

As shown in \figref{fig:supp-domain-radar}, including every main-table baseline places the FedDDA--MOSAIC comparison in context. MOSAIC forms the outer contour on all Office31 domains and on three of four OfficeHome domains, while remaining close to the strongest method on Real World. On DomainNet100, its contour largely overlaps those of FedDDA and FedPGP, with improvements on Clipart, Infograph, and Sketch balanced by small regressions on Painting, Quickdraw, and Real. The radar plot therefore supports the same conclusion as \tabref{tab:supp-domain-results}: the gain is a better macro-client trade-off, not uniform dominance on every domain.
\tabref{tab:supp-class-office31}, \tabref{tab:supp-class-officehome}, and \tabref{tab:supp-class-domainnet} report terminal-checkpoint \method accuracy for every class on Office31, OfficeHome, and DomainNet100, respectively. The tables aggregate 93, 260, and 600 class--domain pairs with held-out sample counts so that all classes remain legible.

\begin{table*}[t]
\centering
\small
\setlength{\tabcolsep}{5pt}
\begin{tabular}{llrrr|rrr}
\toprule
& & \multicolumn{3}{c}{Setting 1: one client/domain} & \multicolumn{3}{c}{Setting 2: two clients/domain} \\
Dataset & Domain & FedDDA & MOSAIC & $\Delta$ & FedDDA & MOSAIC & $\Delta$ \\
\midrule
Office31 & Amazon & 89.32 & 90.57 & +1.25 & 88.39 & 90.00 & +1.61 \\
         & Webcam & 97.14 & 100.00 & +2.86 & 95.06 & 98.10 & +3.04 \\
         & DSLR & 98.23 & 100.00 & +1.77 & 98.10 & 96.90 & $-1.20$ \\
         & \textbf{Average} & 94.90 & \textbf{96.86} & +1.96 & 93.85 & \textbf{95.00} & +1.15 \\
\midrule
OfficeHome & Art & 87.07 & 88.60 & +1.53 & 86.69 & 88.25 & +1.56 \\
           & Clipart & 78.67 & 80.70 & +2.03 & 77.98 & 77.05 & $-0.93$ \\
           & Product & 96.32 & 97.40 & +1.08 & 95.27 & 95.05 & $-0.22$ \\
           & Real World & 93.52 & 93.40 & $-0.12$ & 92.96 & 93.60 & +0.64 \\
           & \textbf{Average} & 88.89 & \textbf{90.03} & +1.13 & 88.22 & \textbf{88.49} & +0.27 \\
\midrule
DomainNet100 & Clipart & 89.74 & 90.30 & +0.56 & 88.65 & 89.65 & +1.00 \\
             & Infograph & 77.34 & 78.80 & +1.46 & 76.07 & 78.30 & +2.23 \\
             & Painting & 88.46 & 88.40 & $-0.06$ & 86.67 & 87.15 & +0.48 \\
             & Quickdraw & 63.67 & 63.30 & $-0.37$ & 61.76 & 59.45 & $-2.31$ \\
             & Real & 94.14 & 93.80 & $-0.34$ & 93.47 & 93.65 & +0.18 \\
             & Sketch & 88.06 & 88.10 & +0.04 & 86.80 & 85.95 & $-0.85$ \\
             & \textbf{Average} & 83.57 & \textbf{83.78} & +0.21 & 82.24 & \textbf{82.36} & +0.12 \\
\bottomrule
\end{tabular}
\caption{Complete domain-level FedDDA and MOSAIC top-1 accuracy (\%) under the two client-partition protocols in the main paper. Average is the macro-client accuracy; because every domain contains the same number of clients, it is also the unweighted mean of the domain-level accuracies. $\Delta$ is MOSAIC minus FedDDA.}
\pdfanchor{tab:supp-domain-results}\label{tab:supp-domain-results}
\end{table*}

\input{generated/supp_class_diagnostic_tables.tex}
% \FloatBarrier

\end{document}

%% file: generated/supp_class_diagnostic_tables.tex
% Generated by analysis/generate_mosaic_final_class_tables.py
\begin{table*}[t]
\centering
\scriptsize
\setlength{\tabcolsep}{2.0pt}
\renewcommand{\arraystretch}{1.10}
\begin{tabular}{lr@{\hspace{5pt}}lr}
\toprule
Class & Acc. & Class & Acc. \\
\midrule
back\_pack & 96.0 & mouse & 100.0 \\
bike & 100.0 & mug & 95.8 \\
bike\_helmet & 100.0 & paper\_notebook & 89.3 \\
bookcase & 91.3 & pen & 96.2 \\
bottle & 69.2 & phone & 95.8 \\
calculator & 100.0 & printer & 92.3 \\
desk\_chair & 100.0 & projector & 96.9 \\
desk\_lamp & 96.0 & punchers & 75.9 \\
desktop\_computer & 96.3 & ring\_binder & 85.7 \\
file\_cabinet & 95.7 & ruler & 61.1 \\
headphones & 96.4 & scissors & 100.0 \\
keyboard & 92.6 & speaker & 100.0 \\
laptop\_computer & 96.9 & stapler & 80.0 \\
letter\_tray & 100.0 & tape\_dispenser & 82.1 \\
mobile\_phone & 100.0 & trash\_can & 83.3 \\
monitor & 100.0 &  &  \\
\bottomrule
\end{tabular}
\caption{Class-level \method accuracy (\%) on Office31 at the saved terminal checkpoint.}
\pdfanchor{tab:supp-class-office31}\label{tab:supp-class-office31}
\end{table*}

\begin{table*}[t]
\centering
\scriptsize
\setlength{\tabcolsep}{2.0pt}
\renewcommand{\arraystretch}{1.10}
\begin{tabular}{lr@{\hspace{5pt}}lr@{\hspace{5pt}}lr}
\toprule
Class & Acc. & Class & Acc. & Class & Acc. \\
\midrule
Alarm\_Clock & 100.0 & Flowers & 100.0 & Postit\_Notes & 83.9 \\
Backpack & 93.0 & Folder & 83.3 & Printer & 98.1 \\
Batteries & 91.5 & Fork & 83.8 & Push\_Pin & 90.9 \\
Bed & 98.0 & Glasses & 100.0 & Radio & 88.9 \\
Bike & 96.8 & Hammer & 85.7 & Refrigerator & 91.3 \\
Bottle & 88.2 & Helmet & 100.0 & Ruler & 89.5 \\
Bucket & 91.7 & Kettle & 93.6 & Scissors & 95.0 \\
Calculator & 100.0 & Keyboard & 86.7 & Screwdriver & 81.6 \\
Calendar & 94.0 & Knives & 93.8 & Shelf & 97.5 \\
Candles & 95.5 & Lamp\_Shade & 75.0 & Sink & 85.4 \\
Chair & 94.5 & Laptop & 85.5 & Sneakers & 98.2 \\
Clipboards & 84.6 & Marker & 83.3 & Soda & 78.6 \\
Computer & 75.4 & Monitor & 68.3 & Speaker & 86.2 \\
Couch & 87.0 & Mop & 84.2 & Spoon & 82.5 \\
Curtains & 93.2 & Mouse & 96.0 & Table & 84.1 \\
Desk\_Lamp & 85.7 & Mug & 98.0 & Telephone & 91.1 \\
Drill & 88.9 & Notebook & 98.1 & ToothBrush & 88.6 \\
Eraser & 78.6 & Oven & 80.6 & Toys & 95.5 \\
Exit\_Sign & 97.7 & Pan & 91.2 & Trash\_Can & 90.0 \\
Fan & 100.0 & Paper\_Clip & 83.3 & TV & 81.5 \\
File\_Cabinet & 97.2 & Pen & 62.0 & Webcam & 92.3 \\
Flipflops & 100.0 & Pencil & 71.7 &  &  \\
\bottomrule
\end{tabular}
\caption{Class-level \method accuracy (\%) on OfficeHome at the saved terminal checkpoint.}
\pdfanchor{tab:supp-class-officehome}\label{tab:supp-class-officehome}
\end{table*}

\begin{table*}[t]
\centering
\scriptsize
\setlength{\tabcolsep}{2.0pt}
\renewcommand{\arraystretch}{1.10}
\resizebox{0.7\textwidth}{!}{%
\begin{tabular}{lr@{\hspace{5pt}}lr@{\hspace{5pt}}lr@{\hspace{5pt}}lr}
\toprule
Class & Acc. & Class & Acc. & Class & Acc. & Class & Acc. \\
\midrule
aircraft\_carrier & 70.7 & bed & 80.3 & cactus & 87.6 & coffee\_cup & 77.2 \\
airplane & 89.3 & bee & 91.4 & cake & 59.6 & compass & 79.3 \\
alarm\_clock & 73.0 & belt & 61.6 & calculator & 87.6 & computer & 81.4 \\
ambulance & 85.2 & bench & 72.4 & calendar & 74.6 & cookie & 80.0 \\
angel & 90.2 & bicycle & 94.2 & camel & 88.8 & cooler & 64.1 \\
animal\_migration & 90.4 & binoculars & 81.8 & camera & 92.2 & couch & 73.5 \\
ant & 73.9 & bird & 78.8 & camouflage & 71.5 & cow & 88.7 \\
anvil & 76.3 & birthday\_cake & 69.4 & campfire & 85.3 & crab & 86.9 \\
apple & 89.4 & blackberry & 76.6 & candle & 84.0 & crayon & 81.3 \\
arm & 80.1 & blueberry & 77.4 & cannon & 77.1 & crocodile & 87.2 \\
asparagus & 84.9 & book & 82.2 & canoe & 82.9 & crown & 92.8 \\
axe & 74.0 & boomerang & 79.2 & car & 89.1 & cruise\_ship & 90.2 \\
backpack & 85.7 & bottlecap & 66.4 & carrot & 85.9 & cup & 67.4 \\
banana & 88.9 & bowtie & 86.1 & castle & 90.4 & diamond & 91.0 \\
bandage & 76.3 & bracelet & 77.1 & cat & 91.8 & dishwasher & 74.7 \\
barn & 89.5 & brain & 89.0 & ceiling\_fan & 82.7 & diving\_board & 76.9 \\
baseball & 75.7 & bread & 81.9 & cell\_phone & 76.2 & dog & 83.7 \\
baseball\_bat & 75.0 & bridge & 81.7 & cello & 92.8 & dolphin & 87.2 \\
basket & 78.4 & broccoli & 86.8 & chair & 66.3 & donut & 94.7 \\
basketball & 80.8 & broom & 83.0 & chandelier & 86.3 & door & 88.7 \\
bat & 71.0 & bucket & 72.1 & church & 84.3 & dragon & 85.5 \\
bathtub & 78.4 & bulldozer & 88.5 & circle & 85.6 & dresser & 75.0 \\
beach & 83.3 & bus & 92.9 & clarinet & 82.8 & drill & 73.3 \\
bear & 86.0 & bush & 55.3 & clock & 63.7 & drums & 85.9 \\
beard & 87.3 & butterfly & 92.4 & cloud & 78.5 & duck & 89.9 \\
\bottomrule
\end{tabular}
}
\caption{Class-level \method accuracy (\%) on DomainNet100 at the saved terminal checkpoint.}
\pdfanchor{tab:supp-class-domainnet}\label{tab:supp-class-domainnet}
\end{table*}